\documentclass[10pt,twocolumn,letterpaper]{article}

\usepackage{cvpr}
\usepackage{times}
\usepackage{epsfig}
\usepackage{graphicx}
\usepackage{amsmath}
\usepackage{amssymb}
\usepackage{cite}
\usepackage{epsfig,latexsym}
\usepackage{float}
\usepackage{indentfirst}
\usepackage{array}
\usepackage{amsmath}
\usepackage{amssymb}
\usepackage{times}
\usepackage{psfrag}

\usepackage{subfigure}
\usepackage{textcomp}
\usepackage{multirow}
\usepackage{multicol}
\usepackage{stfloats}
\usepackage{url}
\usepackage{subfigure}
\usepackage{multirow}
\usepackage{booktabs}
\usepackage{comment}
\usepackage{threeparttable}
\usepackage{booktabs} % For formal tables
\usepackage{subfigure}
\usepackage{url}
\usepackage{multirow}
\usepackage{multicol}
\usepackage{graphicx}
\usepackage{enumerate}
\usepackage{bm}
\usepackage{caption}
\usepackage{color}
\usepackage[linesnumbered,ruled]{algorithm2e}
\usepackage{diagbox}

\newtheorem{theorem}{$\mathbf{Theorem}$}
\newtheorem{lemma}[theorem]{$\mathbf{Lemma}$}

\usepackage[breaklinks=true,bookmarks=false]{hyperref}

\cvprfinalcopy % *** Uncomment this line for the final submission

\def\cvprPaperID{****} % *** Enter the CVPR Paper ID here

\begin{document}

%%%%%%%%% TITLE
\title{Saliency Prediction on Omnidirectional Images with Generative Adversarial Imitation Learning}

\author{Mai~Xu\\
	School of Electronic and Information Engineering, Beihang University\\
		Beijing, 100191\\
%	{\tt\small firstauthor@i1.org}
	\and
	Li~Yang\\
	School of Electronic and Information Engineering, Beihang University\\
	Beijing, 100191\\
%	{\tt\small secondauthor@i2.org}
}

\maketitle
%\thispagestyle{empty}

%%%%%%%%% ABSTRACT
\begin{abstract}
		When watching omnidirectional images (ODIs), subjects can access different viewports by moving their heads. Therefore, it is necessary to predict subjects' head fixations on ODIs. Inspired by generative adversarial imitation learning (GAIL), this paper proposes a novel approach to predict saliency of head fixations on ODIs, named SalGAIL. First, we establish a dataset for attention on ODIs (AOI). In contrast to traditional datasets, our AOI dataset is large-scale, which contains the head fixations of 30 subjects viewing 600 ODIs. Next, we mine our AOI dataset and determine three findings: (1) The consistency of head fixations are consistent among subjects, and it grows alongside the increased subject number; (2) The head fixations exist with a front center bias (FCB); and (3) The magnitude of head movement is similar across subjects. According to these findings, our SalGAIL approach applies deep reinforcement learning (DRL) to predict the head fixations of one subject, in which GAIL learns the reward of DRL, rather than the traditional human-designed reward. Then, multi-stream DRL is developed to yield the head fixations of different subjects, and the saliency map of an ODI is generated via convoluting predicted head fixations. Finally, experiments validate the effectiveness of our approach in predicting saliency maps of ODIs, significantly better than 10 state-of-the-art approaches.
\end{abstract}
%%%%%%%%% BODY TEXT
\section{Introduction}

In recent years, omnidirectional images (ODIs) have become increasingly popular, along with the rapid development of virtual reality (VR). Different from the traditional 2D images, ODIs provide an immersive and interactive VR viewing experience. Moreover, they enable spherical stimuli, meaning that the range of 360$^\circ\times$180$^\circ$ can be accessible to subjects through the head-mounted display (HMD). In other words, humans can freely move their heads to change viewports for viewing attractive regions. Hence, head fixations play a vital role in modeling visual attention on ODIs. Accordingly, it is necessary to predict head fixations, which can be widely used in many applications of ODIs, e.g., compression \cite{de2017video}, rendering \cite{Stengle2016display} and visual quality assessment \cite{Gaddam2016tiling}.

Most recently, there have emerged several works \cite{LEBRETON2018GBVS360, ZHU2018the, BATTISTI2018a, LING2018a} on predicting the saliency maps of head fixations on ODIs.
For example, Lebreton \textsl{et al.} \cite{LEBRETON2018GBVS360} developed two new models for saliency prediction of head fixations on ODIs, that are based on the traditional 2D saliency prediction models: Boolean Map based Saliency model (BMS) \cite{zhang2016bms} and Graph-Based Visual Saliency (GBVS) \cite{harel2007graph}. Thus, these models are called ``BMS360'' and ``GBVS360'', respectively.
Zhu \textsl{et al.} \cite{ZHU2018the} proposed a multi-plane projection method to predict head fixations on omnidirectional scene, in which several blocks are generated to simulate viewports.
Then, the low-level features (spatial frequency, orientation and color) and high-level semantic features (car and person) are extracted in each block, which are further fused and mapped into the overall saliency map.
The above works can be seen as heuristic methods, as their features for predicting head fixations are hand-crafted.
In fact, the great success of deep learning has boosted the development of saliency prediction on 2D images, which is a closely related area of head fixation prediction on ODIs.
However, none of the existing saliency prediction approaches for ODIs is based on deep learning; therefore, their performance is fair.

Further complication are that the existing head fixation datasets for ODIs are all small-scale collections, which can hardly be used to train the deep learning models. Specifically, \cite{Rai2017a} is the first ODI dataset with human attention, which is composed of the head and eye fixations of 63 subjects on 98 ODIs. In addition, both the head and eye fixations of 169 subjects on 22 ODIs are available in the dataset of \cite{Sitzmann2018saliency}.
In \cite{upenik2017simple}, the dataset has 104 ODIs viewed by 40 subjects.
However, only the head fixations data are available in \cite{upenik2017simple} without any eye fixations data.
Moreover, the dataset of \cite{hu2017head} includes the attention data of 27 subjects who were asked to view a total of 70 different ODIs in the VR environment.

In this paper, we establish a large-scale dataset for attention on ODIs (called the AOI dataset), which is comprised by head and eye fixations data of 30 subjects viewing 600 ODIs.
Note that the 600 ODIs of our AOI dataset are diverse in both the resolution and content.
By mining our dataset, we find that high consistency exists for head fixations among subjects when viewing ODIs. Besides, we have some additional findings. (1) The distribution of head fixations are variant between individual subjects; however, the consistency of head fixations tends to increase and converge when the number of subjects increases. (2) The front center bias (FCB) characteristic exists for the head fixations. (3) The magnitude of head movement (HM) is similar across all subjects over all ODIs in our AOI dataset.
Based on the above findings, this paper proposes a generative adversarial imitation learning (GAIL) based approach for saliency prediction of head fixations on ODIs, which is called SalGAIL.

\begin{table*}[!tb]
	\centering
	\caption{Basic properties of the existing ODI/ODV datasets.}
	\resizebox{1\linewidth}{!}{
		\begin{tabular}{|ccccccccc|}
			\hline
			& Dataset & Scene & Images/videos & Subjects & Resolution & Durations (s)  & Ground-truth recorded & HMD/Eye-tracker  \\
			\hline
			\multirow{5}[2]{*}{IMAGE} & Rai \textsl{et al.} \cite{Rai2017a} & Static & 98 & 40 & $\leq$18,332$\times$9,166 & 25 & Head and eye fixations & Oculus Rift DK2/SMI Eye-tracker \\
			& Sitzmann \textsl{et al.} \cite{Sitzmann2018saliency} & Static & 22 & 169 & $\leq$ 8,192$\times$4,096  & 30 & Head and eye fixations & Oculus Rift DK2/Tobii EyeX Eye-tracker \\
			& Upenik \textsl{et al.} \cite{upenik2017simple} & Static & 104 & 40 & 1,334$\times$750  & - & Head fixations & MERGE VR Goggles\footnote{The introduction of this VR device is available on https://mergevr.com/.} plus iPhone 6 \\
			& Hu \textsl{et al.} \cite{hu2017head} & Static & 70 & 27 & 640$\times$480 & 10 & Head fixations & Google Cardboard  \\
			& Abreu \textsl{et al.} \cite{de2017look} & Static & 21 & 32 & 4,096$\times$2,048 & 10/20 & Head fixations & Oculus Rift DK2 \\
			\hline
			\multirow{9}[2]{*}{VIDEO} & Yu \textsl{et al.} \cite{Yu2015a}  & Dynamic & 10 & 10 & $\leq$ 6,144$\times$3,072 & 10 & Head fixations & Oculus Rift DK2 \\
			& Lo \textsl{et al.} \cite{lo2017360}  & Dynamic & 10 & 50 & 4,096$\times$2,048 & 60 & Head fixations & Oculus Rift DK2 \\
			& Xu \textsl{et al.} \cite{xu2018gaze}  & Dynamic & 208 & 31 & 4,096$\times$2,048 & 20-60 & Eye fixations & HTC Vive/aGlass Eye-tracker \\
			& Zhang \textsl{et al.} \cite{zhang2018saliency}  & Dynamic & 104 & 27 &  - & 20-60 & Head and eye fixations & HTC Vive/aGlass Eye-tracker \\
			& Ozcinar \textsl{et al.} \cite{ozcinar2018visual}  & Dynamic & 6 & 17 & $\leq$ 8,192$\times$4,096 & 10 & Head fixations & WebVR \cite{webVR} \\
			& Corbillon  \textsl{et al.} \cite{Corbillon2017360Degree}  & Dynamic & 7 & 59 & 3,840$\times$2,048 & 70 & Head fixations & Razer OSVR HDK2 HMD \\
			& Xu \textsl{et al.} \cite{xu2018predicting}  & Dynamic & 76 & 58 & $\leq$ 8,192$\times$4,096 & 10-80 & Head and eye fixations & HTC Vive/aGlass Eye-tracker \\
			& Deep 360 Pilot  \cite{Hu2017e}  & Dynamic & 342 & 5 & - & - & Annotate salient object in panorama & Without using HMD \\
			& David  \cite{David2018a}  & Dynamic & 19 & 57 & $\leq$3,840$\times$1,920 & 20 & Head and eye fixations & Oculus Rift DK2/SMI Eye-tracker \\
			\hline
			\multirow{1}[2]{*}{} & \textbf{Our dataset } & Static & 600 & 30 & $\leq$24,028$\times$12,014 & 22 & Head and eye fixations & HTC Vive/aGlass Eye-tracker \\
			
			\hline
		\end{tabular}
	}%
	\label{tab:ODI datset}
\end{table*}

Specifically, our SalGAIL approach predicts the head fixations through a deep reinforcement learning (DRL) model. In the DRL model, we regard the directions of head trajectories as the \textit{actions} of the DRL model and take the viewed omnidirectional content as the \textit{observation} of the \textit{environment}. As such, the DRL model can be learned to predict the head fixations of one subject on an ODI. Then, multi-stream DRL is used to generate the head fixations of different subjects, and the predicted head fixations are convoluted to generate the saliency map of the input ODI. However, different from the traditional DRL tasks, the \textit{reward} is intractable to be obtained and quantified in our task for saliency prediction on ODIs. Instead, we propose to learn \textit{reward} by imitating the head trajectories of subjects in the training stage. This strategy benefits from the most recent success of GAIL \cite{ho2016generative, li2017infogail}.

In brief, the main contributions of this paper are summarized as follows.

\begin{itemize}
	\item
	We establish a large-scale AOI dataset with several findings about human attention on ODIs.
	\item
	We propose a multi-stream DRL model to predict head fixations on ODIs.
	\item
	We apply GAIL to learn the \textit{reward} of our DRL model by imitating the head trajectories of subjects.
\end{itemize}

%	This paper is organized as follows.
%	Section II reviews the related works.
%	Section III presents the established AOI dataset.
%	In Section IV, we analyze our AOI dataset to obtain some findings as the preliminary of our SalGAIL approach.
%	In Section V, we propose our SalGAIL approach.
%	Section VI reports the experimental results to verify the effectiveness of our SalGAIL approach, and Section VII concludes this paper.
%		

\begin{figure}[!tb]
	\begin{center}
		\includegraphics[width=.75\linewidth]{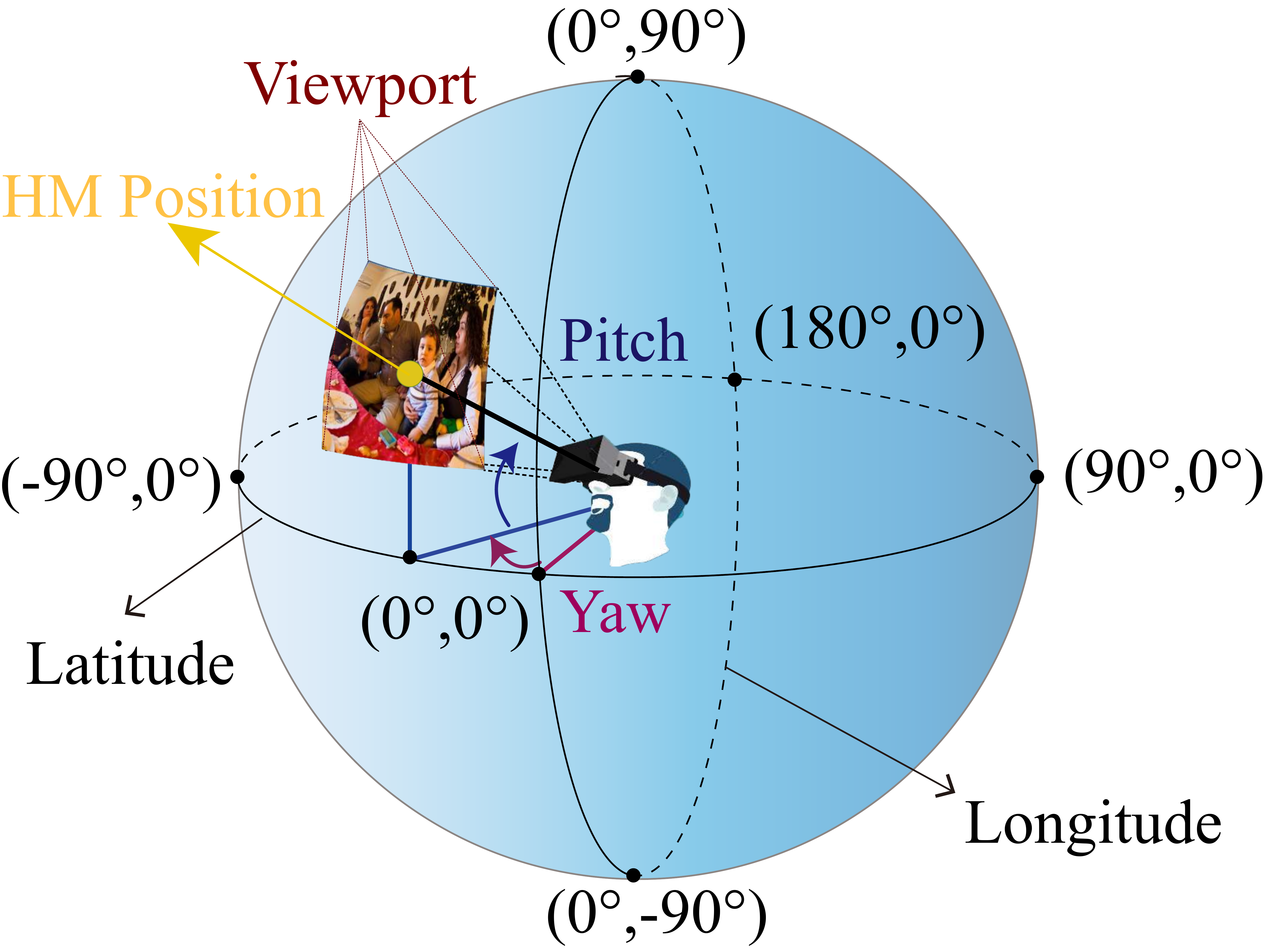}
	\end{center}
	%\vspace{-1em}
	\caption{An example of HM in the sphere. The latitude and longitude of the HM position, i.e., the center of a viewport, are only dependent on the angles of pitch and yaw, respectively.}
	\label{fig:coord}
\end{figure}
%-------------------------------------------------------------------------
\section{Related Works}
\label{sec:related-works}

In this section, we review the approaches and datasets for saliency prediction on ODIs.

\subsection{Saliency prediction approaches for ODIs}
\label{relate_w sub1}
\textbf{Saliency prediction on 2D images.}
The past two decades have witnessed extensive works on saliency prediction for 2D images, such as \cite{zhang2016bms}, \cite{cheng2015global, Judd2009learning, goferman2012context, yang2017top, Kanan2009SUNTS, Ramanishka2017top} . In the task of image saliency prediction, many effective spatial features have been proposed in predicting human attention with either a top-down or a bottom-up strategy. Specifically, Itti \textsl{et al.} \cite{Itti1998a} considered low-level features at multiple scales and combined them to form the saliency map of an image. Harel \textsl{et al.} \cite{harel2007graph} introduced a graph-based visual saliency (GBVS) model that defines
Markov chains over various image maps, and treated the equilibrium distribution over map locations as activation and saliency values. Considering top-down image semantics, Judd \textsl{et al.} \cite{Judd2009learning}  proposed a saliency model based on low-, middle- and high-level image features. Moreover, Borji \textsl{et al.} \cite{borji2012boosting} proposed combining low-level features of the bottom-up models with top-down cognitive visual features, and then learning a direct mapping from those features to eye fixations.
Inspired by deep learning, deep neural networks (DNNs) have been successfully used to predict image saliency in an end-to-end manner, such as \cite{cornia2016a, pan2016shallow, pan2017salgan, huang2015salicon, Kruthiventi2015deepfix, simonyan2013deep, Zhao_2015_CVPR, kmmerer2017understanding}. Specifically, a DNN-based structure was proposed in Deepfix \cite{Kruthiventi2015deepfix} to learn a multiscale semantic representation for image saliency. Moreover, saliency in context (SALICON) was proposed in \cite{huang2015salicon}, which fine-tunes the existing DNNs with an effective saliency-related loss function. In \cite{kmmerer2017understanding}, a readout architecture was proposed for image saliency prediction, in which both low-level and DNN features are considered.

\begin{figure*}
	\centering
	\includegraphics[width=0.9\linewidth]{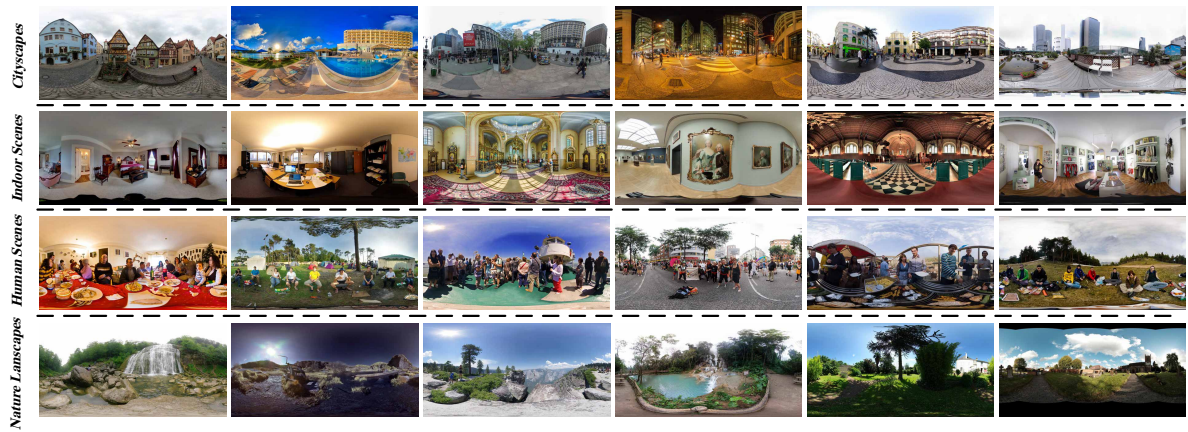}
	\caption{Some examples of ODIs in our AOI dataset.}
	\label{fig:dabase_show}
\end{figure*}

\textbf{Saliency prediction of eye fixations on ODIs.}
Although saliency prediction has been well developed for 2D images, there are only a few approaches for predicting the saliency maps of ODIs.
Different from 2D images, the saliency of ODIs refers to two forms: head fixations and eye fixations.
Most of the existing saliency prediction approaches for ODIs focus on eye fixations, including \cite{BATTISTI2018a, ZHU2018the, LING2018a, LEBRETON2018GBVS360, startsev2018360, Monroy2017salnet360, Reina2017SaltiNet}.
In particular,  Battisti \textsl{et al.} \cite{BATTISTI2018a} presented a saliency model for predicting the saliency maps of eye fixations on ODIs, which is based on the combination of low-level and semantic features. Startsev \textsl{et al.} \cite{startsev2018360} proposed a new saliency prediction approach by considering projection distortions, equator bias and vertical border effects, for predicting saliency of eye fixations on ODIs. In addition, Ling \textsl{et al.} \cite{LING2018a} took human color perception into account and proposed a model using color dictionary-based sparse representation for ODI saliency prediction.
Besides, DNNs have also been successfully applied \cite{Monroy2017salnet360, Reina2017SaltiNet} for saliency prediction on ODIs.
``SalNet360''  \cite{Monroy2017salnet360} was proposed to fine-tune traditional CNN models of 2D saliency prediction for the task of ODI saliency prediction.
Additionally, ``SaltiNet'' \cite{Reina2017SaltiNet} was developed to train a DNN model for eye fixation prediction on ODIs, which is based on a temporal-aware representation of saliency information called saliency volume.

\textbf{Saliency prediction of head fixations on ODIs.} In particular, there are relatively few works \cite{LEBRETON2018GBVS360, de2017look, ZHU2018the} to predict the saliency maps of head fixations for ODIs. Zhu \textsl{et al.} \cite{ZHU2018the} employed a method of multiview projection to generate the saliency maps of head fixations on ODIs. In their work, an ODI is first projected into multiview blocks to simulate viewports. Then, both bottom-up and top-down features of all blocks are extracted and fused to generate the final saliency map of head fixations.
In addition, \cite{de2017look} proposed adding the center bias of human attention into the saliency maps of ODIs through a postprocessing method.
Lebreton \textsl{et al.} \cite{LEBRETON2018GBVS360} developed new models for saliency prediction on ODI, called ``BMS360'' and ``GBVS360'', which are based on the traditional 2D saliency prediction models, boolean map based saliency model (BMS) \cite{zhang2016bms} and GBVS \cite{harel2007graph}. More specifically, ``BMS360'' applied multiple fusion saliency (FMS) \cite{de2017look} to remove the border constraints. In ``GBVS360'', the input ODI in equirectangular format is projected into several rectilinear images, corresponding to different viewports, and then feature extraction is performed according to each rectilinear image. Finally, the resulting feature maps are back-projected to the equirectangular domain to yield the saliency map.
In contrast to ODIs, more works \cite{xu2018predicting, Hu2017e, xu2018gaze, zhang2018saliency} have been proposed for predicting head fixations on omnidirectional videos (ODVs).
For example, Xu \textsl{et al.} \cite{xu2018predicting}  proposed a DRL approach for saliency prediction of head fixations on ODVs.
Additionally, Zhang \textsl{et al.} \cite{zhang2018saliency} presented a spherical CNN-based scheme for saliency prediction of ODVs. In the following, we overview the existing datasets for attention modeling on ODIs/ODVs.

\subsection{Attention datasets for ODIs}
To learn saliency models on  ODIs/ODVs, datasets with head fixations and eye fixations are urgently required.
Along with saliency prediction approaches, several ODIs/ODVs datasets have been recently established to collect the head fixation/eye fixation data of subjects when viewing omnidirectional scenes.
Table \ref{tab:ODI datset} summarizes the basic properties of these datasets.
To the best of our knowledge, Salient360 (Rai \textsl{et al.} \cite{Rai2017a}) and Saliency in VR (Sitzmann \textsl{et al.} \cite{Sitzmann2018saliency}) have been widely used in the recent ODI saliency prediction works.
These datasets are reviewed in more details as follows.

\textbf{Salient360} (Rai \textsl{et al.} \cite{Rai2017a}) is one of the earliest ODI datasets for saliency prediction. It contains 98 stimuli, which mainly include indoor, outdoor and people scenes. For each ODI, at least 40 subjects were asked to view the stimuli with free head movement in the range of 360$^\circ\times$180$^\circ$.
The maximum resolutions of these stimuli are 18,332$\times$9,166.
Each ODI was presented for 25 seconds with an identical initialized viewport for all subjects.
Then, the eye fixations and head fixations were recorded.
Finally, the ground truth saliency maps of head fixations and eye fixations were both converted into the equirectangular format.

\begin{figure*}
	\centering
	%\captionsetup{justification=centering,margin=2cm}
	\includegraphics[width=0.9\linewidth]{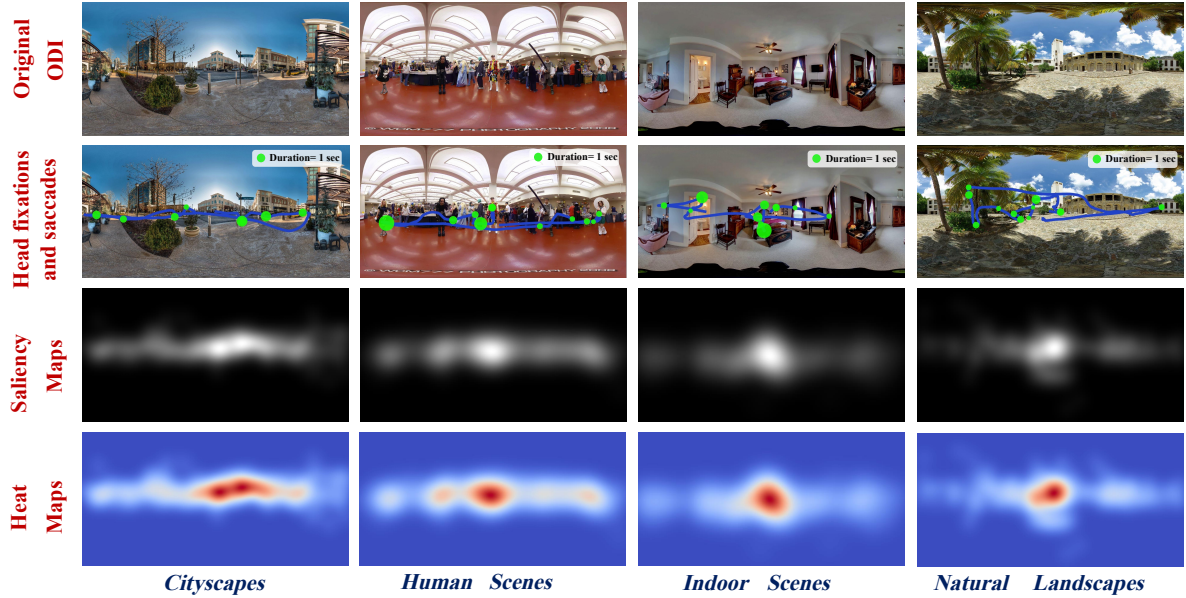}
	\caption{Examples for head fixations and saccades as well as saliency maps (in the form of gray maps and heat maps).}
	\label{fig:salmapn}
\end{figure*}

\textbf{Saliency in VR} (Sitzmann \textsl{et al.} \cite{Sitzmann2018saliency}) is also a public dataset that records 1,980 trajectories of head fixations and eye fixations, obtained from 169 subjects viewing 22 static ODIs.
In their experiment, the data of head fixations and eye fixations were captured using an HMD in both standing (called \textit{VR} \textit{standing} ) and seated (called \textit{VR seated}) conditions.
In addition, \cite{Sitzmann2018saliency} also collected the data of observing the same scenes through a desktop monitor, called \textit{desktop} condition.
The dataset offered the ground truth saliency maps of head fixations and eye fixations using three different projections from sphere to plane, i.e., equirectangular, cube map and patch-based projection.

The attention dataset can benefit the saliency prediction approaches for ODIs/ODVs.
In particular, the deep learning approaches require large-scale data for training the DNN models.
Unfortunately, as shown in Table \ref{tab:ODI datset}, the existing datasets lack sufficient data, especially for ODIs.
Therefore, we establish a large-scale dataset for saliency prediction on ODIs, namely the AOI dataset. The details about our dataset are discussed in Section \ref{sec:AOI_dataset}.

\section{Dataset}

\label{sec:AOI_dataset}
\subsection{Data collection}
\textbf{Stimuli.} First, we collected 600 ODIs from Flickr \cite{Flickr2018}, the resolution of which ranges from 4,000$\times$2,000 to 24,028$\times$12,014. Each ODI was downloaded in the equirectangular format and at the maximum resolution. Note that all 600 ODIs were available under the creative commons copyright. To enrich the diversity of the content in our dataset, four categories of ODIs were collected including \textit{cityscapes, natural landscapes, indoor scenes and human scenes}. Figure \ref{fig:dabase_show} shows some examples for each category of ODIs in our dataset.

\textbf{Equipment.} We obtained the HM and eye movement (EM) data of the subjects through the HTC vive and aGlass. Here, the HTC vive is used as an HMD to view ODIs. The HM data can be captured by the HTC vive, while the aGlass device is able to capture the EM data within FoV. Note that the aGlass device is embedded in the HTC vive. When the subjects viewed ODIs, the ``virtual desktop'' was used to display all images, and meanwhile the software of \cite{Li2018Bridge} was applied to record both HM and EM data. Note that the whole HM data along with the time stamps form the head trajectory of viewing an ODI, from which we can extract the head fixations.
Similarly, the eye fixations can also be obtained from the EM data.

\begin{figure}
	\centering
	\includegraphics[width=0.91\linewidth]{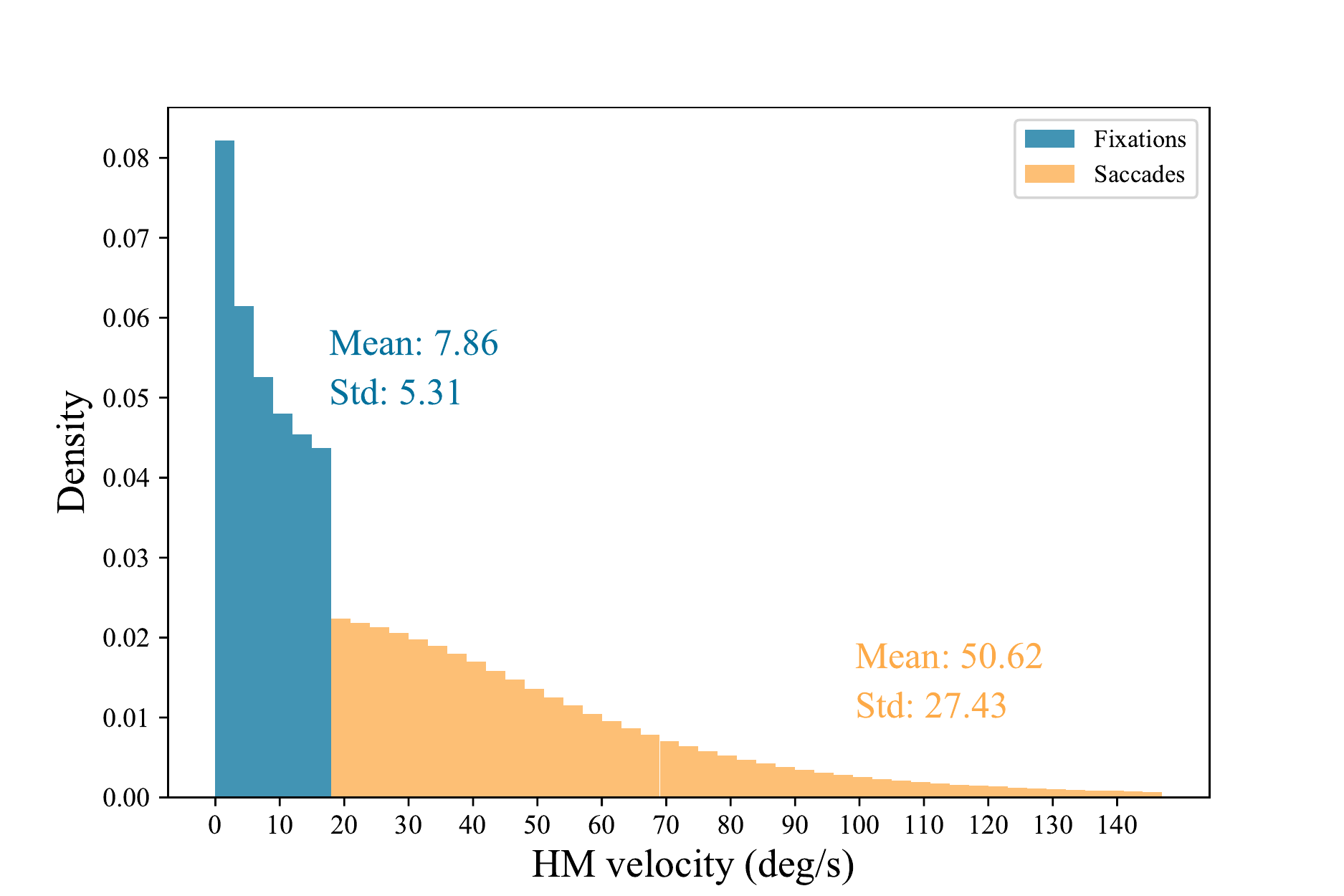}
	\caption{Histogram distribution of the velocities of fixations (blue) and saccades (orange).}
	\label{fig:vel_new}
	\vspace{-2.1em}
\end{figure}
\textbf{Subjects.} There were in total 30 subjects (19 females and 11 males) involved in our experiment, and their ages ranged from 18 to 30, with an average of 21 years old. Note that all subjects have normal or corrected-to-normal eyesight\footnote{The device of aGlass can also be used to correct eyesight.}. Before viewing ODIs, a simple training session was conducted to familiarize the subjects with the HTC vive.
Furthermore, the procedure of the experiment was explained to all subjects. Finally, the subjects underwent the experiment of viewing ODIs with the following procedure.

\textbf{Procedure.}  The 600 ODIs were randomly and equally divided into 2 equal groups. Two groups of ODIs were viewed by each subject on different days to avoid the fatigue. According to the Salient360 dataset \cite{Rai2017a}, the duration was set to 22 seconds for viewing each ODI. After viewing each image, we inserted a gray ODI with a red dot located at longitude = $0^\circ$ and latitude = $0^\circ$, and the subjects pressed to enter the next ODI once they fixated on the red dot. Consequently,  when the subjects viewed the next ODI, their HM and EM can be re-initialized to the center of the corresponding equirectangular image.
Note that the subjects were allowed to have a rest when they felt fatigue.
When viewing ODIs, the subjects wearing the HTC vive were asked to sit in a comfortable swivel chair, allowing them to rotate 360$^\circ$ freely. As such, all panoramic regions in the image can be easily accessed.

\textbf{Raw data.} Then, the raw HM data are recorded in the following format. Note that our AOI dataset does not process the EM data, as this paper only focuses on predicting head fixations on ODIs.
In fact, the HM data of a subject at one ODI can be represented by a vector: [Time stamp, HM pitch, HM yaw].	
%	\\
%	\begin{center}
%			\vspace{-1.0em}
%			[Time stamp, HM pitch, HM yaw].
%			\vspace{-1.0em}
%	\end{center}
Specifically, the above vector is composed of the time stamp and HM position. (1) Time stamp: The interval time between two neighboring sample points of HM are recorded and represented in milliseconds for each ODI.
(2) HM position: Two elements are related to the HM position, including 2 Euler angles: the angles of pitch and yaw. As shown in Figure \ref{fig:coord}, the location of the viewport can be represented in latitude and longitude, corresponding to the angles of pitch and yaw, respectively.

\subsection{Data processing}\label{sec_data}
Given the above raw HM data, we need to distinguish head fixations and saccades. In this paper, we mainly focus on predicting the saliency maps of head fixations. However, our dataset can also be used to predict HM saccades of HM for the future work.
Our algorithm for distinguishing head fixations and saccades is presented as follows.

\textbf{Head fixations and saccades.} When viewing ODIs, one subject may move his/her head along with saccades, and then fix on the regions that are attractive to him/her, seen as head fixations.
Next, we focus on extracting head fixations and saccades from the HM data. First, the velocity of HM is measured through the \textit{orthodromic distance} \cite{ORH} between two successive HM data divided by the corresponding time stamp. Mathematically, the HM velocity $v_i$ at the \textit{i}-th sample can be denoted as
\begin{equation}
v_i = \frac{d_i}{\Delta T},
\end{equation}
where $\Delta T$ is the duration of the time stamp, and $d_i$ is the \textit{orthodromic distance} between the (\textit{i}-1)-th and \textit{i}-th samples. Here, $d_i$ is defined as follows:
\begin{equation}
d_i = r \cdot \Delta \sigma_i,
\end{equation}
where $\Delta \sigma_i$ is the \textit{spherical distance}:
\begin{equation}
\Delta \sigma_i = 2\arcsin\sqrt{\sin^{2}(\dfrac{\Delta\psi_i}{2}) + \cos\psi_i \cdot \cos\psi_{i-1} \cdot \sin^{2}(\dfrac{\Delta\theta_i}{2})}.
\end{equation}
In addition, \textit{r} is the radius of the omnidirectional sphere. For the \textit{i}-th HM sample, $\Delta \psi_i$ is the difference in latitude; $\Delta \theta_i$ is the difference in longitude; and $\psi_{i-1}$ and $\psi_i$ are latitudes of two successive samples.

Then, using the \textit{velocity-threshold identification algorithm (I-VT)}  \cite{Salvucci2000Identifying}, we separate the head fixations and saccades based on the sample-to-sample velocities of HM. In this paper, we follow \cite{upenik2017simple} to set the velocity threshold to be 18 degrees/second. In other words, if the velocity of an HM sample is below 18 degrees/second, it belongs to the head fixations; otherwise it belongs to saccades.
Figure \ref{fig:vel_new} shows the histograms of head fixations and saccades, calculated over all HM data of 600 ODIs in our dataset.
Since this paper mainly focuses on predicting head fixations, all saccades (HM with the speed above the threshold) are discarded prior to the further analysis. After this process, we obtain the head fixations for our dataset. Figure \ref{fig:salmapn}-(Second row) shows a raw head trajectory of one subject when viewing an ODI, which is composed of saccades and head fixations.

\textbf{Saliency maps of head fixations.} For obtaining the 2D saliency maps of head fixations, we apply equirectangular projection to process the sphere-format data according to \cite{Zhou2017a}. In equirectangular projection, the yaw and pitch of the $i$-th head fixation on the sphere coordinate (in degrees) are mapped to a 2D pixel in the equirectangular image, i.e., a head fixation denoted by ($x_i, y_i$) in the equirectangular coordinate. Here, the origin (0, 0)  of the equirectangular coordinate is located at the lower left corner of the ODI. Then, ($x_i, y_i$) can be obtained by

\begin{equation}
\begin{aligned}
x_i = (\frac{\theta_i}{360} + \frac{1}{2}) \times  W \nonumber\text{,}
\end{aligned}
\end{equation}
\begin{equation}
\begin{aligned}
y_i = (\frac{\psi_i}{180} + \frac{1}{2}) \times H \text{.}
\end{aligned}
\end{equation}
For the $i$-th head fixation, $\theta_i$ and $\psi_i$ are its yaw and pitch, respectively; $W$ and $H$ denote the width and height of the equirectangular image, in the form of pixel numbers.

Then, the head fixations of all subject are convolved with a Gaussian kernel to generate the saliency map for each ODI. According to \cite{Iwasaki1986relation,David2018a}, a 2D Gaussian kernel with 1.5$^\circ$ visual angle centered at the head fixation is used in this paper. The kernel that locates at head fixation ($x_i, y_i$) can be represented as:
\begin{equation}
\label{gau_kel}
\begin{aligned}
G(x_i, y_i) = \frac{1}{2\pi\sigma^2}\cdot e^{-\frac{x_i^2 + y_i^2}{2\sigma^2}},
\end{aligned}
\end{equation}
where the standard deviation is $\sigma = \frac{W}{2\sqrt{2 \rm ln2}}$ with $W$ a constant value of 90 pixels \cite{Rajashekar2005StatisticalAA}.  Figure \ref{fig:salmapn} shows some examples of the original equirectandular images and saliency maps.

\begin{figure}
	\begin{minipage}{0.48\linewidth}
		\centering
		\includegraphics[width=1.05\linewidth]{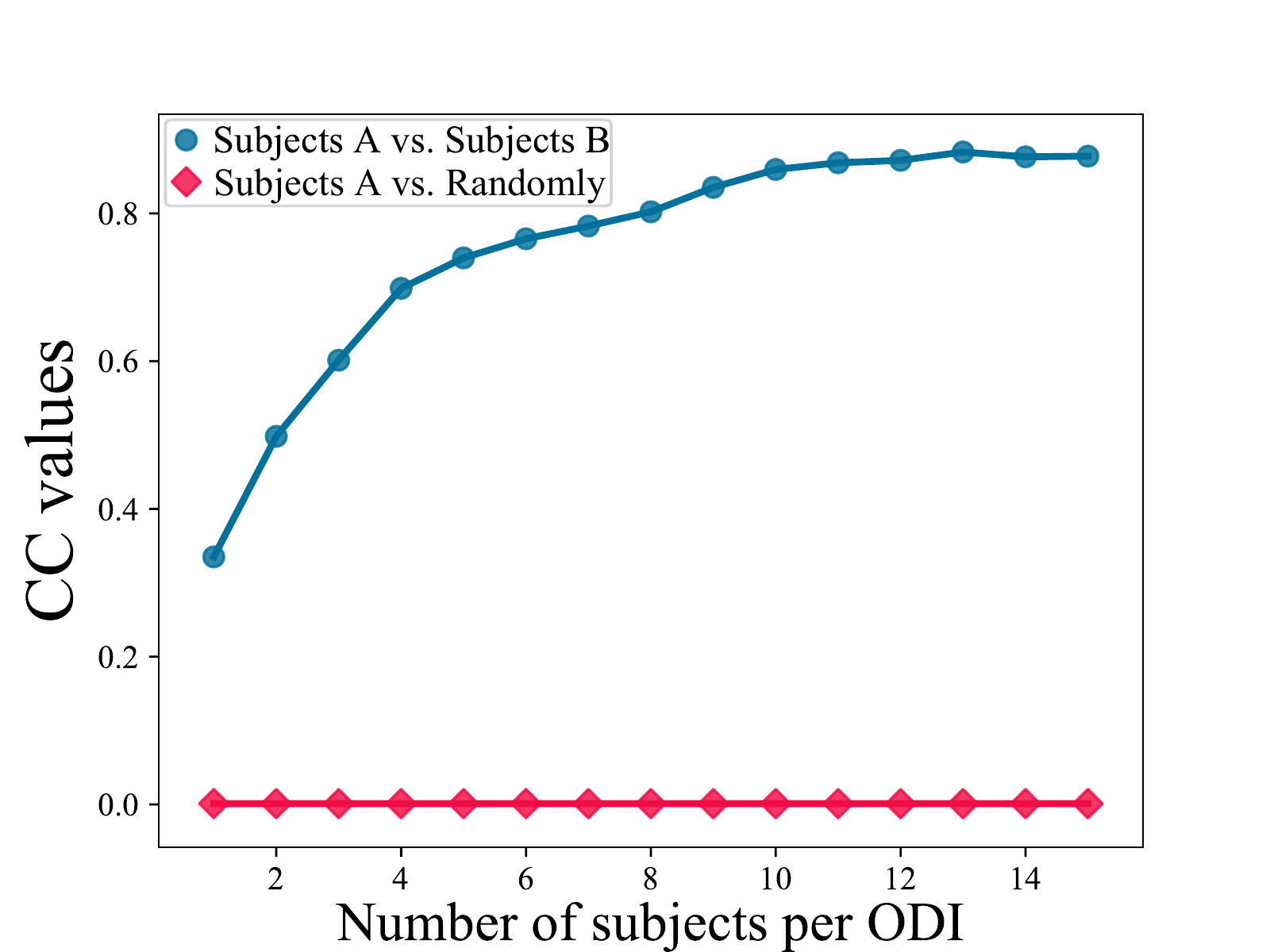}\\
		\centerline{\footnotesize{(a) }}\medskip
	\end{minipage}
	\hfill
	\begin{minipage}{0.48\linewidth}
		\centering
		\includegraphics[width=1.05\linewidth]{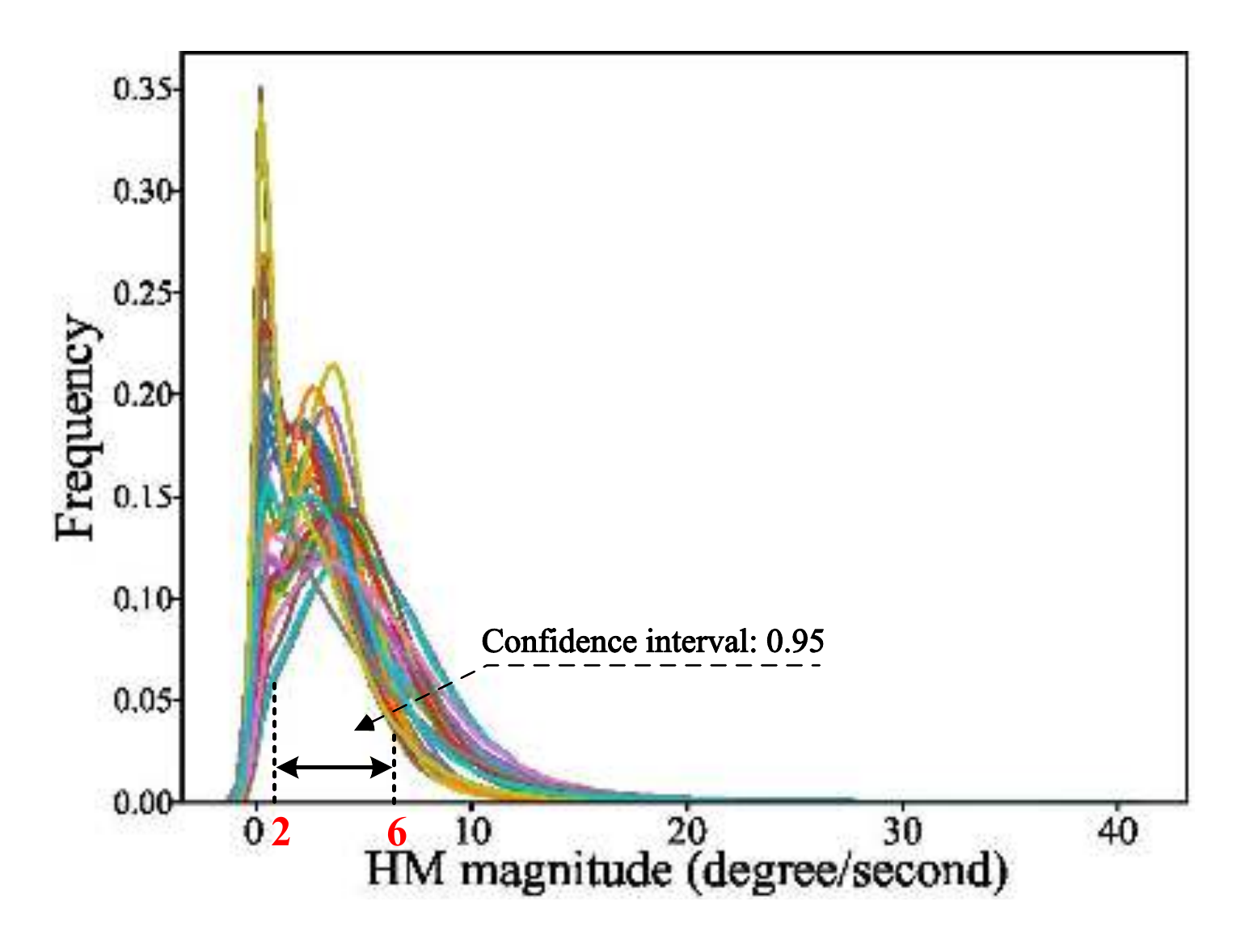}\\
		\centerline{\footnotesize{(b) }}\medskip
	\end{minipage}
	\hfill
	
	\caption{\footnotesize{(a): Average CC values alongside increased numbers of subjects per ODI over our AOI dataset; The two curves denote CC values between A and B, randomly sampled maps, respectively. (b): Probability distribution for the HM magnitude between adjacent HM across 30 subjects, over 600 ODIs of our AOI dataset. Each curve stands for the distribution of one subject.}}
	\label{fig:finding1_3}
\end{figure}

\begin{figure*}
	\begin{minipage}{0.32\linewidth}
		\centering
		\includegraphics[width=1.05\linewidth]{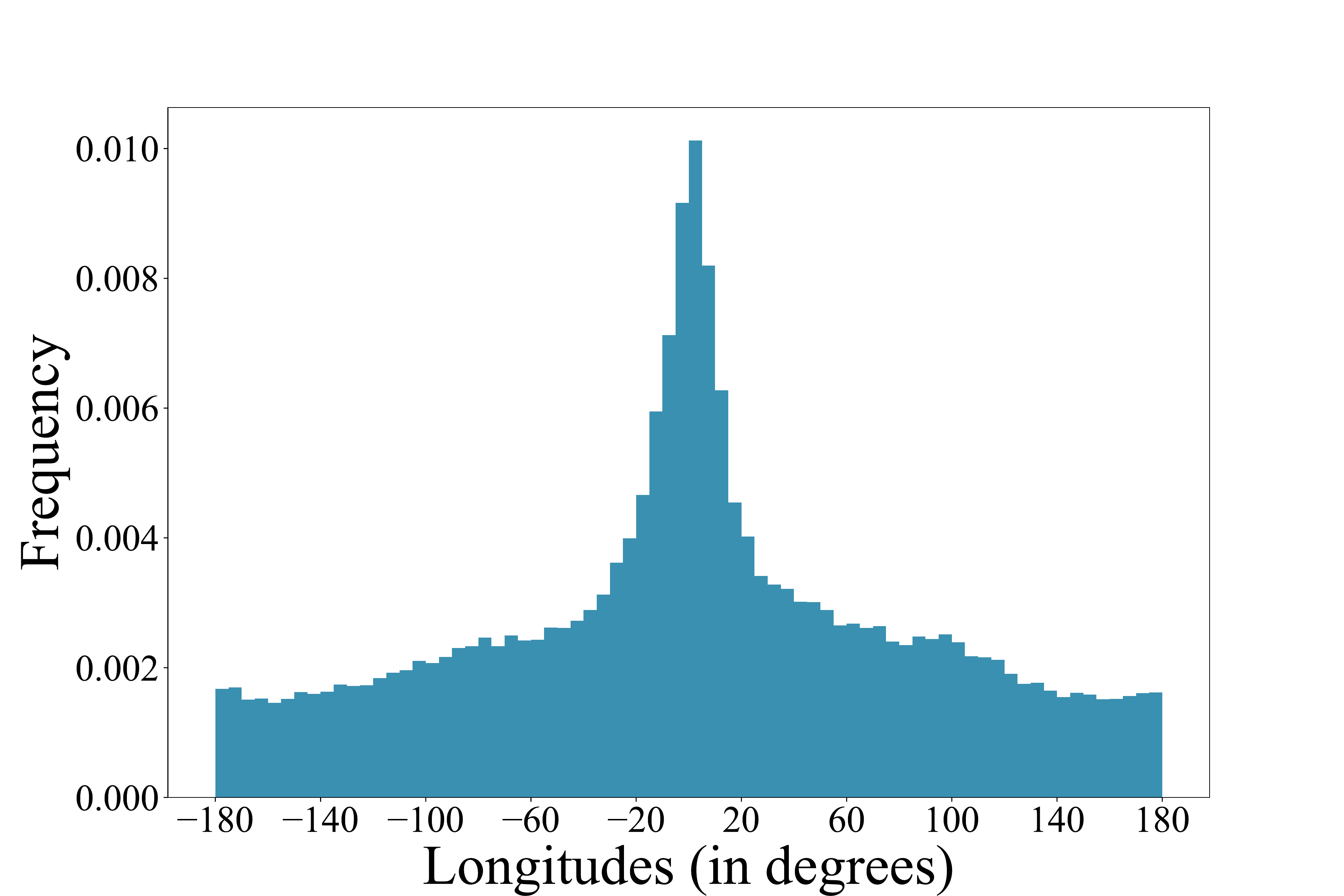}\\
		\centerline{\footnotesize{(a) Frequency distribution alongside longitude}}\medskip
	\end{minipage}
	\hfill
	\begin{minipage}{0.32\linewidth}
		\centering
		\includegraphics[width=1.05\linewidth]{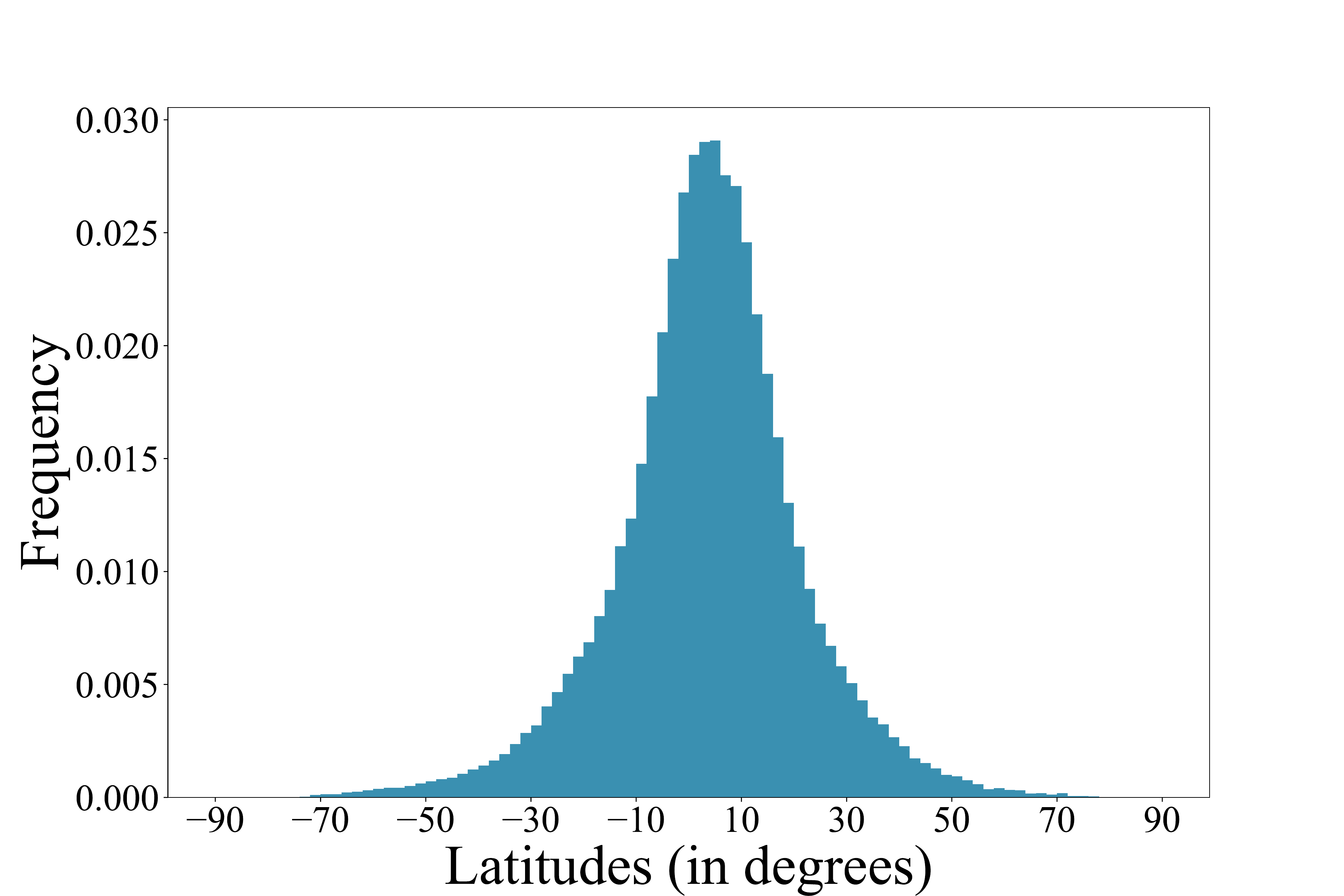}\\
		\centerline{\footnotesize{(b) Frequency distribution alongside latitude}}\medskip
	\end{minipage}
	\hfill
	\begin{minipage}{0.32\linewidth}
		\centering
		\includegraphics[width=1.01\linewidth]{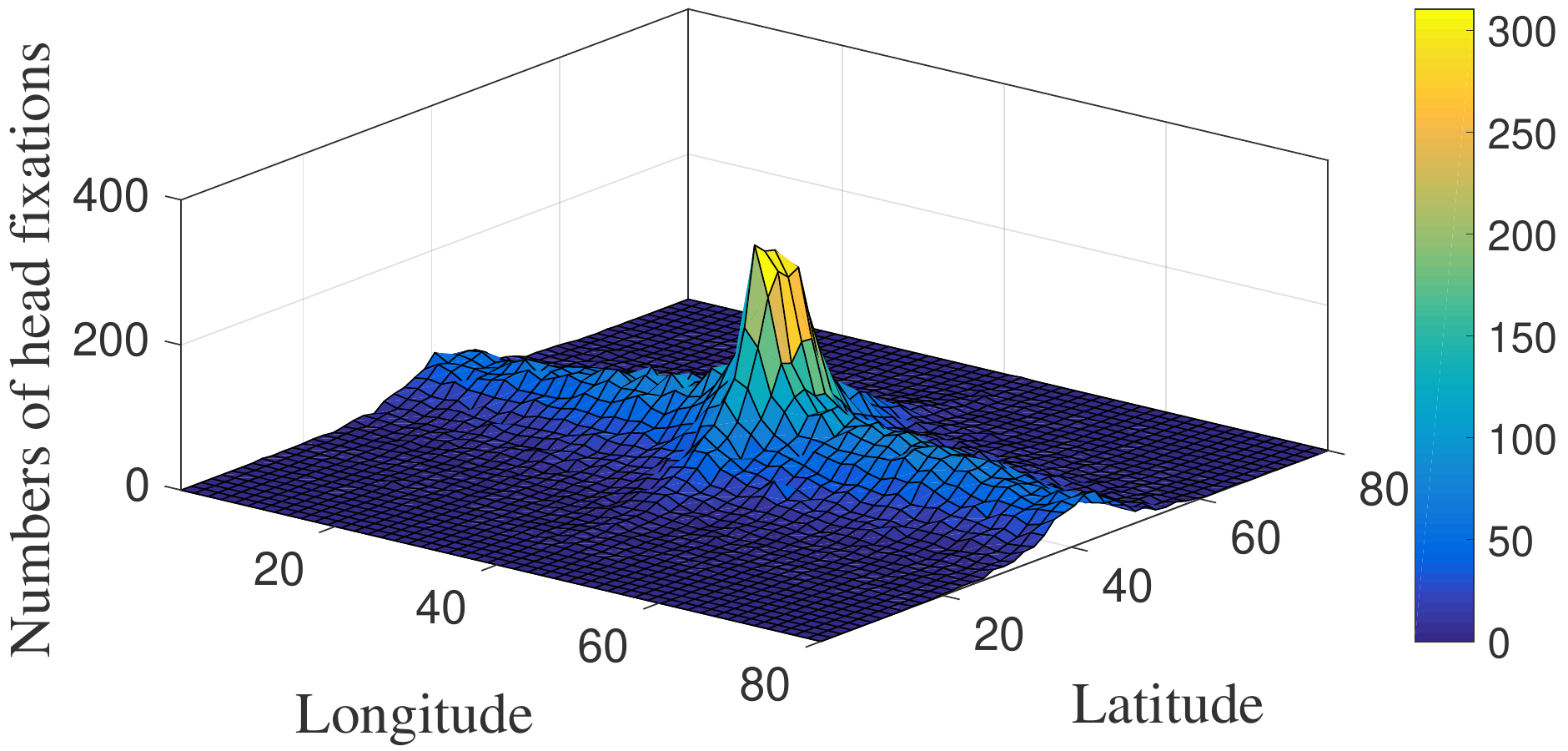}\\
		\centerline{\footnotesize{(c) Frequency distribution at different regions}}\medskip
	\end{minipage}
	\caption{\footnotesize{(a): Numbers of head fixations alongside longitude over (from -180$^\circ$ to 180$^\circ$), (b): Numbers of head fixations alongside latitude (from -90$^\circ$ to 90$^\circ$),  (c): Numbers of head fixations at different omnidirectional regions, calculated over all 30 subjects and all ODIs in our AOI dataset.}}
	\label{fig:centerbias}
\end{figure*}

\section{Dataset analysis}

In this section, we mine our dataset to investigate human behavior in viewing ODIs. Specifically, we have the following findings.

\textit{Finding 1}: \textit{The distribution of head fixations are variant between individual subjects; however, the consistency of head fixations among subjects increases and converges when the number of subjects increases.}

\textit{Analysis}:
We analyze the consistency of the head fixation distribution across subjects when the number of subjects increases.
To this end, we randomly divide the subjects into two equal groups, denoted as $A$ and $B$, and the number of subjects in these two groups progressively increases from 1 to 15.
For each ODI in our dataset, the saliency maps of $A$ and $B$ are generated by convolving with the  2D Gaussian kernel (see \eqref{gau_kel}) over the corresponding head fixations, which are denoted as $\mathbf{S}_A$ and $\mathbf{S}_B$, respectively.
Then, the consistency of head fixations between two groups is measured by calculating the linear correlation coefficient (CC) of saliency maps between $\mathbf{S}_A$ and $\mathbf{S}_B$.
Figure \ref{fig:finding1_3}-(a) shows the average CC values between $\mathbf{S}_A$ and $\mathbf{S}_B$ along with the increased number of subjects.
We also plot the CC values between saliency map $\mathbf{S}_A$ and the saliency map of randomly generated head fixations (with the same number as group $A$).
Note that in Figure \ref{fig:finding1_3}-(a), the CC values are calculated and averaged over all ODIs, after randomly dividing $A$ and $B$ 20 times.
We can see from this figure that the CC value is 0.335 when the subject number is 1 in each group.
This result indicates that the head fixations are variant between two subjects, despite a certain consistency that exists (larger than that between the subject and the randomly generated head fixations).
In addition, the CC value increases and converges along with the increased subject number.
Therefore, this completes the analysis of \textit{Finding 1}.

\textit{Finding 2}: \textit{There exists FCB for the head fixations on ODIs. }

\textit{Analysis}: Given all collected fixations in our dataset, we calculate the distribution of their locations along with the longitude and latitude. The results are shown in Figure \ref{fig:centerbias}-(a) and (b), respectively. We can see from this figure that the head fixations tend to be attracted by the regions near $0^\circ$ longitude (i.e., the front region) and $0^\circ$ latitude (i.e., the equator). Hence, high probability exists that the head fixations fall into the front center region. In other words, the FCB holds for the head fixations on ODIs.
In addition, Figure \ref{fig:centerbias}-(c) counts the numbers of head fixations in different omnidirectional regions over our dataset. In this figure, the full equirectangular region of 360$^\circ$ $\times$ 180$^\circ$ panorama is equally segmented to 4.5$^\circ$ $\times$ 2.25$^\circ$ grids. Then, the numbers of head fixations of all 30 subjects are counted in each grid. As observed in Figure \ref{fig:centerbias}-(c), head fixations are more likely to be attracted by the equator (i.e., the latitude is close to $0^\circ$), especially the center of the equator (i.e., the longitude is also close to $0^\circ$). Again, this observation verifies that the FCB exists for the head fixations in our dataset. Therefore, the analysis of \textit{Finding 2} is substantiated.

\textit{Finding 3}: \textit{The HM magnitude is similar across all subjects over all ODIs in our AOI dataset.}

\textit{Analysis}: For each subject, we calculate the magnitude between two HM positions of two adjacent samples through the \textit{spherical distance}. Figure \ref{fig:finding1_3}-(b) shows the distributions of HM magnitudes for all subjects in our AOI dataset, and in this figure each curve stands for the distribution of one subject. As can be seen in this figure, the distributions of the HM magnitudes are similar among subjects. In particular, most of the HM magnitude values locate at the range of $2^\circ \sim 6^\circ$ for almost all subjects (confidence interval: 95\%). Consequently, there exists similarity for the HM magnitude across all subjects when viewing ODIs.
This completes the validation of \textit{Finding 3}.

\section{SalGAIL approach}
\subsection{Framework}
\label{Framework of SalGail}
\begin{figure}
	\centering
	\includegraphics[width=1\linewidth]{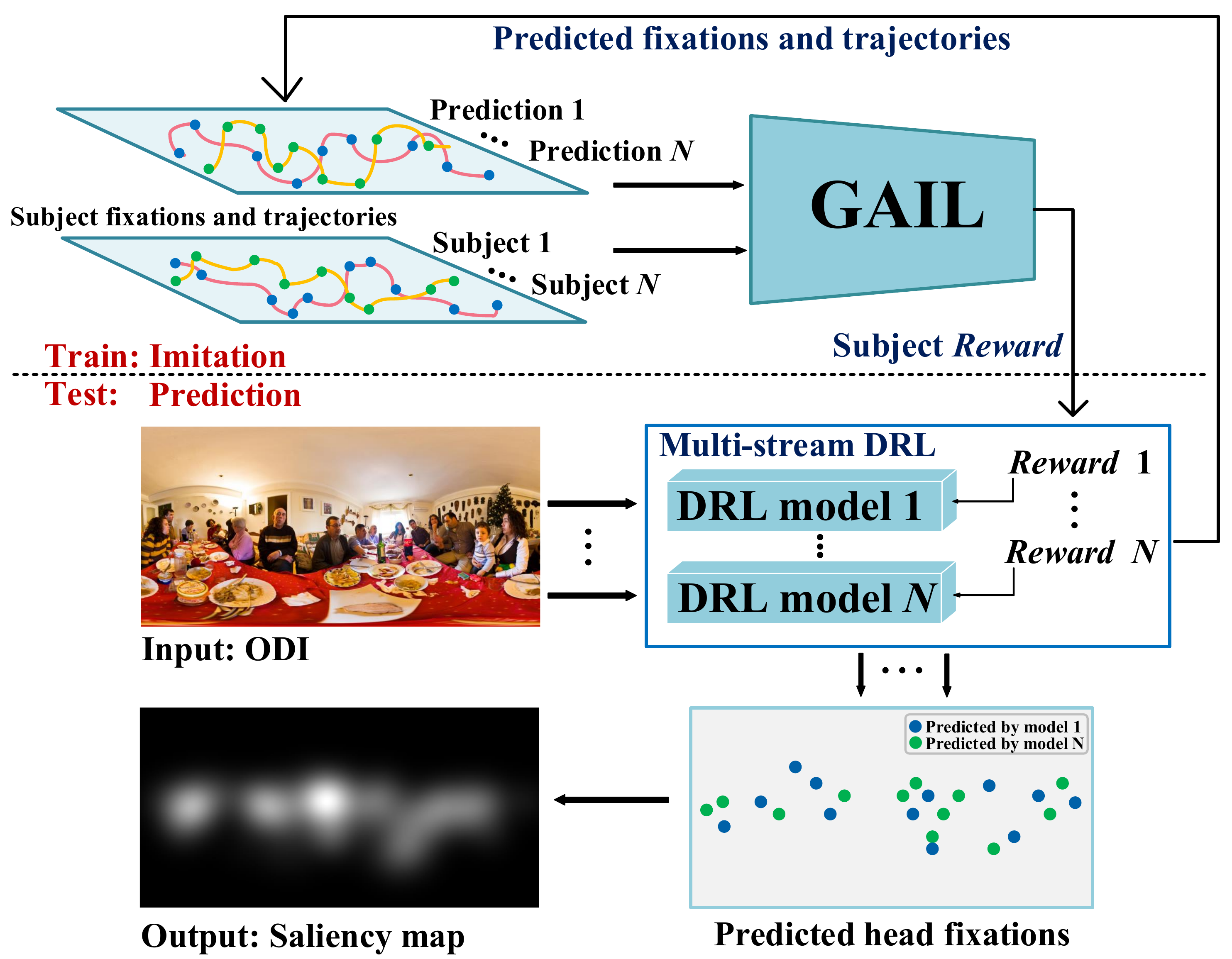}
	\caption{Framework of the SalGAIL approach.}
	\label{fig:cvprfignew}
	\vspace{-1.7em}
\end{figure}

In this section, we present our SalGAIL approach that aims to predict head fixations on ODIs in the form of saliency maps.
Figure \ref{fig:cvprfignew} shows the overall framework of the SalGAIL approach.
As seen in this figure, our SalGAIL approach is composed of two stages: training and test.
In the training stage, we propose a GAIL method for learning the \textit{reward} of imitating the head fixations of each subject.
Then, the learned \textit{reward} of each subject is used in the corresponding DRL stream to predict a head trajectory.	
In the test stage, an ODI is input to a multi-stream DRL model, and given the learned \textit{reward}; then, each DRL stream predicts one head trajectory.
Consequently, the predicted head fixations can be obtained from the head trajectories of all DRL streams.
Finally, the predicted head fixations are convoluted to generate the saliency map of the input ODI.

\subsection{Test: multi-stream DRL for saliency prediction}
\label{Test}
\textbf{Problem formulation}. First, we formulate the problem of saliency prediction on the input ODI (denoted by $\mathbf{I}$) as follows.
Assume that there are in total $N$ DRL streams, each of which corresponds to one subject.
Since there are 30 subjects in our AOI dataset, \textit{N} is chosen to be 30 in this paper. Note that the head fixations of 30 subjects have converged to consistency, according to \textit{Finding 1}.
Then, we establish an $N$-stream DRL model to predict the head trajectories of $N$ subjects, in which the $n$-th stream aims to generate the head trajectory of subject \textit{n}, denoted as  $\bm{\tau}^n= \{(\hat{x}_t^{n},\hat{y}_t^{n})\}_{t=1}^T$.
Here, $\hat{x}_t^{n}$ and $\hat{y}_t^{n}$ are the 2D coordinates of the HM position obtained from the $n$-th DRL stream at time step $t$, and $T$ means the total duration of viewing each ODI.
Next, we need to extract the head fixations from all predicted head trajectories  $ \{\bm{\tau}^n\}_{n=1}^N= \{(\hat{x}_t^{n},\hat{y}_t^{n})\}_{t=1,n=1}^{T,N}$.
Let $\{\mathbf{p}_k^n\}_{k=1,n=1}^{K_n,N}$ denote all extracted head fixations, where $\{\mathbf{p}_k^n\}_{k=1}^{K_n}$ is the set of all head fixations from the $n$-th DRL stream, and $K_n$ is the total number of fixations output by this DRL stream.
Then, saliency map $\tilde{\mathbf{S}}$ can be generated for the input ODI, via convoluting all fixations $\{\mathbf{p}_k^n\}_{k=1,n=1}^{K_n,N}$ with a 2D Gaussian kernel (see \eqref{gau_kel}).
Since \textit{Finding 2} reveals that the FCB exists for head fixations, FCB map $\mathbf{C}$ is added into the generated saliency map $\tilde{\mathbf{S}}$ to output the final saliency map $\mathbf{S}$:
\begin{equation}
\label{cb1}
\mathbf{S} = \text{Norm} (\mathbf{C} + \tilde{\mathbf{S}}),
\end{equation}
where Norm$(\cdot)$ is the normalization operation that ensures all saliency values range from 0 to 1.

\textbf{Multi-stream DRL model}.
Now, we focus on the multi-stream DRL model for predicting all head trajectories $\{(\hat{x}_t^{n},\hat{y}_t^{n})\}_{t=1,n=1}^{T,N}$.
In our approach, each multi-stream DRL shares the same framework, but with different \textit{rewards}.
See Section \ref{GAIL} for more details about the \textit{reward} modeling.
Here, we take the $n$-th DRL stream as an example.
Specifically, we define the terms of the DRL stream as follows.

\begin{itemize}
	\item
	\textit{Observation} $\hat{\mathbf{O}}_t^n$ is the viewport at time step $t$ for the $n$-th DRL stream.
	%It is worth mentioning that the size of the viewport is determined by the HMD.
	%At step $\mathit{t}$, the \textit{observation} $\hat{o}_t^n$ is corresponded to the viewport ($110^\circ\times110^\circ$) in VR, which is extracted by making its center locates at HM position $({\hat{x}}_t^{n}, \hat{{y}}_t^{n})$.
	\item
	\textit{Action} $\hat{a}_{t}^n$ at time step $t$ is formulated by the HM from $t-1$ to $t$. Similar to \cite{xu2018predicting}, the \textit{action} space includes 8 discrete directions $\{0^\circ, 45^\circ, 90^\circ, \cdots, 315^\circ\}$.
	Different from \cite{xu2018predicting}, one additional \textit{action} (denoted as $stay$) is added to indicate that the HM is fixed without any change in the viewport. Thus, the \textit{action} space is $\{0: stay, 1: 0^\circ, 2: 45^\circ, 3: 90^\circ, \cdots, 8: 315^\circ\}$.
	\item
	\textit{Policy} $\pi_{\bm{\omega}}^n$ is modeled as the predicted probability distribution over the \textit{actions} of HM across time steps with $\bm{\omega}$ as its parameters.

	\item
	\textit{Reward} $r_t^n$ denotes the \textit{reward} of the \textit{action} made at time step $t$ in the $n$-th DRL stream. In our approach, the \textit{reward} function is learned to imitate the human \textit{actions} of head trajectories, to be described in Section \ref{GAIL}.
	
	\item
	\textit{Environment} $\mathbf{E}$ is composed of the \textit{reward} estimator and viewport extractor, such that the \textit{reward} and \textit{observation} can be obtained for the \textit{agent} in action-making.
	
\end{itemize}

\begin{figure}
	\begin{minipage}{1\linewidth}
		\centering
		\includegraphics[width=1.0\linewidth]{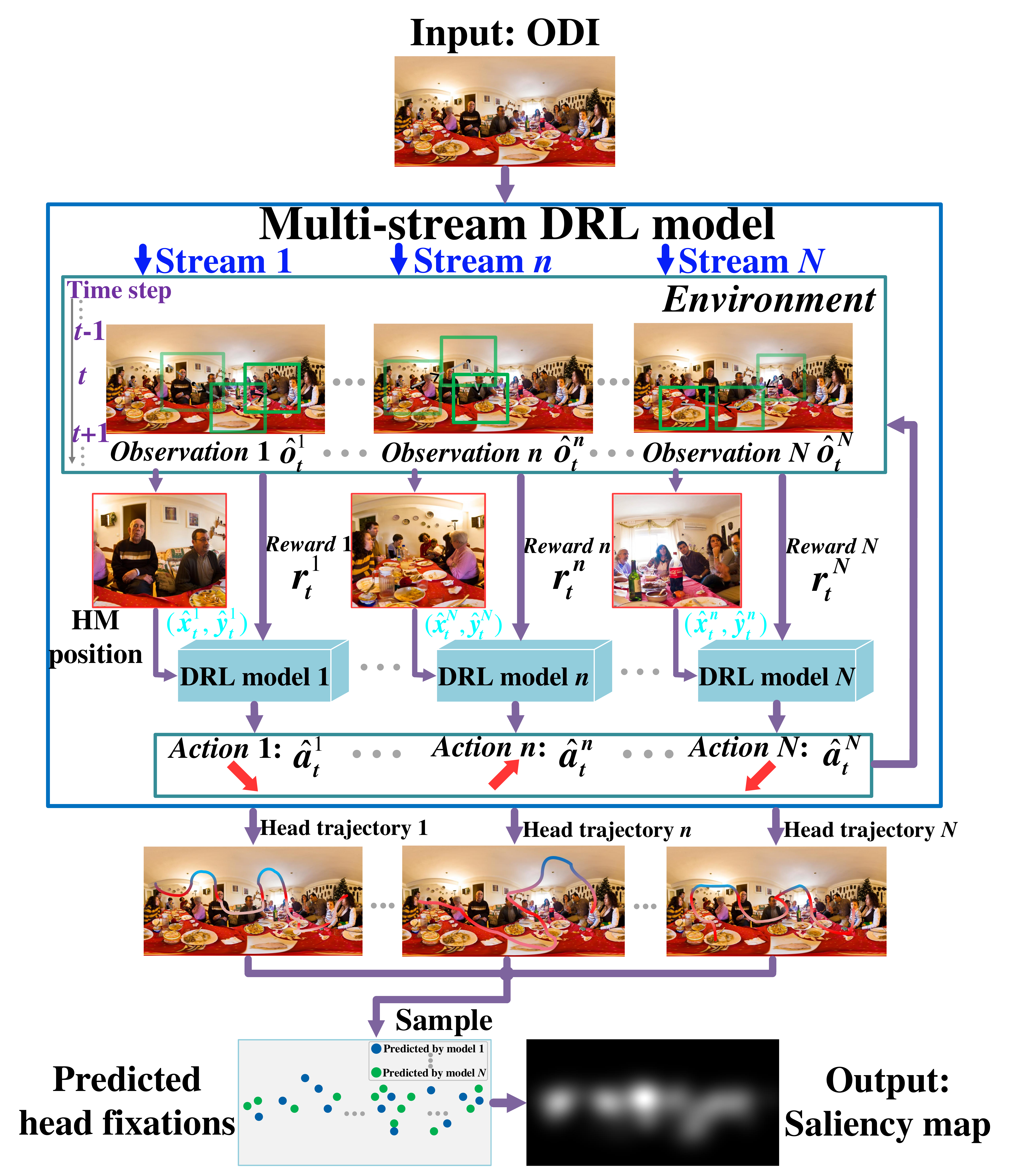}\\
		\centerline{\footnotesize{(a) Framework of our SalGAIL approach}}\medskip
		
	\end{minipage}
	\vfill
	\begin{minipage}{1\linewidth}
		\centering
		\includegraphics[width=1.0\linewidth]{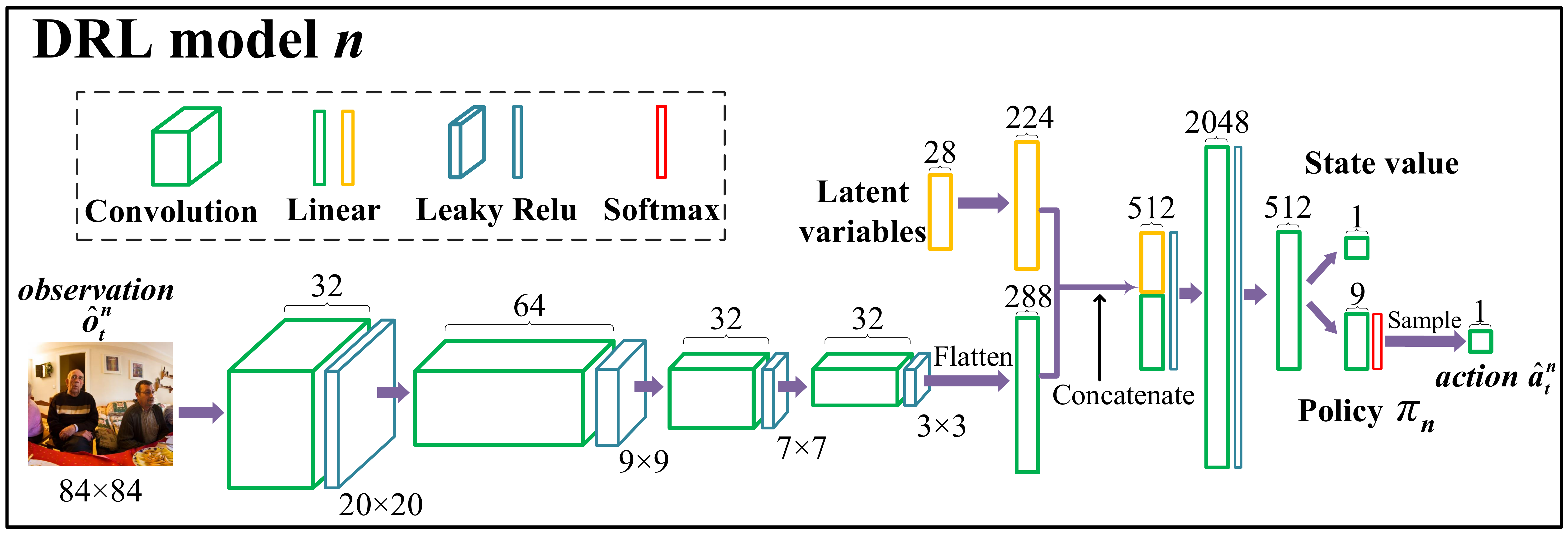}\\
		\centerline{\footnotesize{(b) Structure of CNN in DRL model $n$}}\medskip
	\end{minipage}
	\vfill
	\caption{\footnotesize{(a): Framework of our SalGAIL approach for predicting the saliency maps of head fixations on ODIs. (b): Taking the $n$-th DRL stream for an example, we show the internal structure of the DRL network.}}
	\label{fig:framework_DRL-CNN}
\end{figure}

Given the above terms, the procedure of our multi-stream DRL is summarized in Figure \ref{fig:framework_DRL-CNN}.
Specifically, for the $n$-th DRL stream, \textit{observation} $\hat{\mathbf{O}}_t^n$ at time step $t$ is obtained from the viewport.
That is, the viewport is extracted to make its center locate at the HM position $({\hat{x}}_t^{n}, \hat{{y}}_t^{n})$.
It is worth mentioning that the size of the viewport is determined by the HMD.
Then, the viewport is projected onto the 2D plane with the size of  $84\times84$, as \textit{observation} $\hat{\mathbf{O}}_t^n$.
Subsequently, \textit{observation} $\hat{\mathbf{O}}_t^n$ is input into a CNN (see Figure \ref{fig:framework_DRL-CNN}-(b) for the structure of CNN) to produce a \textit{policy} $\pi_n$, which maximizes \textit{reward} $r_{t}^n$.
Given the \textit{policy}, the \textit{agent}  follows $\epsilon$-\textit{greedy} \cite{Auer2002} to randomly sample an action $\hat{a}_t^{n}$  from $stay$ or 8 discrete directions.
Based on action $\hat{a}_t^{n}$, \textit{environment} $\mathbf{E}$ updates the current HM position with a fixed HM magnitude, from $({\hat{x}}_t^{n}, \hat{{y}}_t^{n})$ to $({\hat{x}}_{t+1}^{n}, \hat{{y}}_{t+1}^{n})$ for the next time step. Here, we set the fixed HM magnitude to be averaged magnitude across all subjects on the ODIs of the training set according to \textit{Finding 3}.
Then, the new \textit{observation} $\hat{\mathbf{O}}_{t+1}^n$ can be obtained upon $({\hat{x}}_{t+1}^{n}, \hat{{y}}_{t+1}^{n})$ for making the \textit{action} at time step $t+1$. The transition of the \textit{observation} is then defined as \textit{T}: $\hat{\mathbf{O}}_{t+1}^n\sim T(\hat{\mathbf{O}}_t^n\mid\hat{a}_t^n)$.

\subsection{Training: GAIL for reward modeling}
%\vspace{-6em}
\label{GAIL}

%\vspace{-3.em}
\begin{algorithm*}
	\footnotesize{				
		\caption{Training stage based on GAIL.}
		\label{alg:GAIL}
		\DontPrintSemicolon
		\SetKw{OR}{or}
		\SetKw{AND}{and}
		\KwIn{The training ODIs; the head trajectories of $N$ subjects $\{\bm{\tau}_{S}^n\}_{n=1}^N$ over the training ODIs.}
		\KwOut{The learned $\textit{policies}$ $\{\pi_{\bm{\omega}}^n\}_{n=1}^N$ of the multi-stream DRL model.}% }
		
		\textbf{Initialize:} Maximum number of training cycles $H$; the number of episodes $I$; the step size of one episode $B$; random initial parameters $\bm{\omega}_0=\{\bm{\omega}_{\pi}^0, \bm{\omega}_{V}^0\}$, $\bm{\phi}_0$ and $\bm{\eta}_0$ for the generator, discriminator and \textit{policy} selector. \;
		
		Run the generator, discriminator and \textit{policy} selector with the initial parameters $\bm{\omega}_0, \bm{\phi}_0$ and $\bm{\eta}_0$, respectively. \;

		%		$\mathit{N} \leftarrow 28 $ (The number of streams in multi-stream DRL model)\;
		%		$\mathit{E} \leftarrow 42 $  (The number of episodes running one ODI) \;
		%		$\mathit{B} \leftarrow 5 $  (Step size in one episode) \;
		\While {$h<H$}
		{
			$h$ $\leftarrow$ $h$+1. \;
			Randomly sample $N$ ODIs into the multi-stream DRL model. \;
			
			Initialize HM positions $\{(\hat{x}_t^n, \hat{y}_t^n)\}_{t=1, n=1}^{N}$ $\leftarrow$ (0, 0), as \textit{observations} $\{\hat{\mathbf{O}}_t^n\}_{t=1, n=1}^{N}$.
			
			\For {$i = 1 \to I$}
			{
				Initialize empty sets $\bm{\chi}$ and $\bm{\chi}_S$ for collecting the  $\textit{observation-action}$ pairs of prediction and subjects, respectively.
				
				\For {$n = 1, 2, \dots, N$}
				{
					Set latent vector $\mathbf{c}_{n}$ in an one-hot form, corresponding to $n$-dimension. \;
					%$\mathbf{c}_{n}\sim\mathit{P}(\mathbf{c}_n)$.\;
					
					Obtain \textit{observation-action} pairs  $\bm{\chi}_{S}^{n}$ (size: $B$) of the $n$-th subject from $\bm{\tau}_{S}^n$: $\bm{\chi}_{S}^{n}\subset\bm{\tau}_{S}^n$.\;
					
					Sample predicted \textit{observation-action} pairs $\bm{\chi}^n$ (size: $B$) under the $n$-th DRL $\textit{policy}$: $\hat{a}_t^n\sim\pi_{\bm{\omega}}^n(\hat{a}_t^n\mid\hat{\mathbf{O}}_t^n, \mathbf{c}_n)$, $\hat{\mathbf{O}}_{t+1}^n\sim T(\hat{\mathbf{O}}_t^n\mid\hat{a}_t^n)$.\;
					
					Calculate the $\mathit{rewards}$ $\{r_t^n\}_{t=(i-1)\cdot B + 1}^{i \cdot B}$ according to \eqref{reward} and deliver the \textit{rewards} into \textit{rollout} for optimization.\;
					
					\If{$\{r_t^n\}_{t=(i-1)\cdot B + 1}^{i \cdot B}$ $\rm$ tend to converg}
					{		
						\textbf {Break} \textbf{from} \textbf{the} \textbf{while} \textbf{loop}.
					}	
					
					Append $\bm{\chi}_{S}^n$ and  $\bm{\chi}^n$ into $\bm{\chi}_{S}$ and $\bm{\chi}$, respectively.\;
					
				}
				%Obtain   $\{r_t^n\}_{t=(i-1)\cdot B + 1}^{i \cdot B}$, which are the \textit{rewards} $\{r_t^n\}_{t=(i-1)\cdot B + 1}^{i \cdot B}$ of multi-stream DRL model at each episode.\;
				
				Update $\bm{\phi}_i$ to $\bm{\phi}_{i+1}$ by ascending with gradients $\Delta_{\bm{\phi}_i}$ in \eqref{phi}.   \;
				
				%				\begin{center}
				%					$\mathbb{E}_{\bm{\chi_S}}[\nabla_{\phi_i}\log\mathit{D}_{\phi_i}(\mathbf{O}, a)] + \mathbb{E}_{\bm{\chi}}[\nabla_{\phi_i}\log(1 - \mathit{D}_{\phi_i}(\hat{\mathbf{\mathbf{O}}}, \hat{a}))]$
				%				\end{center}
				
				Update $\bm{\eta}_i$ to $\bm{\eta}_{i+1}$ by decreasing with gradients $\Delta_{\bm{\eta}_i}$ in \eqref{psi}. \;
				
				%				\begin{center}
				%					-$\lambda_1\mathbb{E}_{\bm{\chi}}[\nabla_{\psi_i}\log\mathit{S}_{\psi_i}(\bm{c}\mid\mathbf{\hat{O}}, \hat{a})] $
				%				\end{center}
				
				\For {$n = 0, 1, 2, \dots, N$}
				{
					Compute $R_t^n$ using $\mathit{rewards}$  $\{r_t^n\}_{t=(i-1)\cdot B + 1}^{i \cdot B}$ from \textit{rollout} according to \eqref{V_estimate}.\;	
					
					Update $\bm{\omega}_V^i$ to $\bm{\omega}_V^{i+1}$ by decreasing with gradients $\Delta_{\bm{\omega}_V^i}$ in \eqref{theta_V}. \;
					
					%					\begin{center}
					%						$\mathbb{E}_{\theta_V^i}[\nabla_{\theta_V^i}(V^\star(\mathbf{\hat{O}}^n, \hat{a}^n)-V_{\theta_V^i}(\mathbf{\hat{O}}^n, \hat{a}^n))^2]$
					%					\end{center}
					
					Update $\bm{\omega}_{\pi}^i$ to $\bm{\omega}_{\pi}^{i+1}$ by increasing with gradients $\Delta_{\bm{\omega}_{\pi}^i}$ in \eqref{theta_pi}.
					
					%					\begin{center}
					%						$\mathbb{E}_{\theta_{\pi}^i}[\log{\pi_{\theta_{\pi}^i}^n}(\hat{a}^n\mid\hat{\mathbf{O}}^n, \bm{c}_n)A_n(\hat{\mathbf{O}}^n, \hat{a}^n)]-\lambda_2\mathit{H}(\pi_{\theta_{\pi}^i}^n)$
					%					\end{center}
					
				}
				Assign parameters $\bm{\omega}_{V}^{i+1}$ and $\bm{\omega}_{\pi}^{i+1}$ to \textit{policies} $\{\pi_{\bm{\omega}}^n\}_{n=1}^{N}$; assign parameters $\bm{\phi}$ and $\bm{\eta}$ to $D_{\bm{\phi}}$ and $S_{\bm{\eta}}$, respectively. \;
				\Return $\{\pi_{\bm{\omega}}^n\}_{n=1}^{N}$.\;
				
			}
		}
	}
\end{algorithm*}

%$r_d$ from the discriminaor $\mathit{D}_{\phi}$ and $\mathit{reward} r_p$ from the posterior approximation $\mathit{Q}_{\psi}$

In this section, we focus on modeling the \textit{reward} for the DRL model of our SalGAIL approach, which is based on GAIL.
Specifically, GAIL is applied to make the predicted head trajectories of our DRL model imitate ground truth head trajectories of subjects. The framework of the training stage can be seen in Figure \ref{fig:train}-(a).
For GAIL, our multi-stream DRL model acts as the generator, outputting the head trajectories.
Then, the discriminator distinguishes whether a head trajectory is a predicted one or the ground truth.
Consequently, the probability that an input head fixation is the ground truth can be obtained from the discriminator, viewed as the \textit{reward} for the DRL model.
Furthermore, we propose a \textit{policy} selector to match the \textit{policy} of one DRL model to the corresponding subject, when maximizing the \textit{reward} in the discriminator.
%Thus, we propose a \textit{policy} selector,  which aims at matching the \textit{policy} of one DRL model to the corresponding subject.
%Note that the \textit{reward} is maximized in the generator to make the predicted head trajectories close to ground truth.
In the following, more details about the generator, discriminator and selector are presented.

%	 As noted in previous section, we want to learn a policy that mimics a behavior provided via a subject demonstration, which can get access to the maximum accuracy in predicting head fixations. Consequently, we extend GAIL \cite{DBLP:journals/corr/HoE16} to enable the learned policy to match subject-like head movement patterns from collected subject demonstration data. As can be seen in Figure \ref{fig:trainframework}.
\begin{itemize}
	\item {Generator.}
	\label{G_net}
	The generator of our SalGAIL approach is the multi-stream DRL model described in Section \ref{Test} , which aims at learning  \textit{policies} to imitate the head trajectories of subjects.
	The input to the generator is the training ODIs.
	Then, the \textit{policy} of one DRL stream can be updated at each episode for optimizing the corresponding \textit{reward}.
	Note that each DRL stream learns one \textit{policy}, corresponding to the head trajectories of a subject.
	Consequently, the generator outputs the predicted head trajectories, as the input to the discriminator. Here, the predicted head trajectories are $\{\bm{\tau}^n\}_{n=1}^N$ correspond to the ground truth head trajectories of subjects $\{\bm{\tau}_{S}^n\}_{n=1}^N$, which are obtained under the \textit{policies} of these subjects: $\bm{\pi}_S$=$\{\pi_S^n\}_{n=1}^N$.
	
	\item {Discriminator.}
	\label{D net}
	The discriminator is a binary classifier determining whether an input head trajectory is the ground truth, which is also used to yield the \textit{reward} of the DRL model in predicting head trajectories. The parameters of the discriminator are updated by distinguishing samples of the head trajectories red in each episode between subjects and the generator. Specifically, the CNN structure of the discriminator can be seen in Figure \ref{fig:train}-(b). The input to the discriminator is the \textit{observations} sampled from the head trajectories concatenated with the corresponding \textit{actions} that are encoded by one-hot vector (9 dimensions). Here, the samples from the head trajectories are the \textit{observation-action} pairs in each episode extracted from $\{\bm{\tau}_{S}^n\}_{n=1}^N$ and $\{\bm{\tau}^n\}_{n=1}^N$, defined as $\bm{\chi}_{S}$ and  $\bm{\chi}$, respectively.
	Hence, the probability of $\bm{\chi}$ being $\bm{\chi}_{S}$ seen as the output of the discriminator is viewed as the \textit{reward} for our DRL model.
	Unfortunately, the original GAIL approach can only learn the \textit{reward} by imitating a single subject \cite{ho2016generative}.
	It is difficult for each DRL stream to make correct \textit{actions} to imitate the head trajectories of multiple subjects.
	Therefore, we propose a \textit{policy} selector that can be used to solve this problem.

	\begin{figure}
		\begin{minipage}{1\linewidth}
			\centering
			\includegraphics[width=1.0\linewidth]{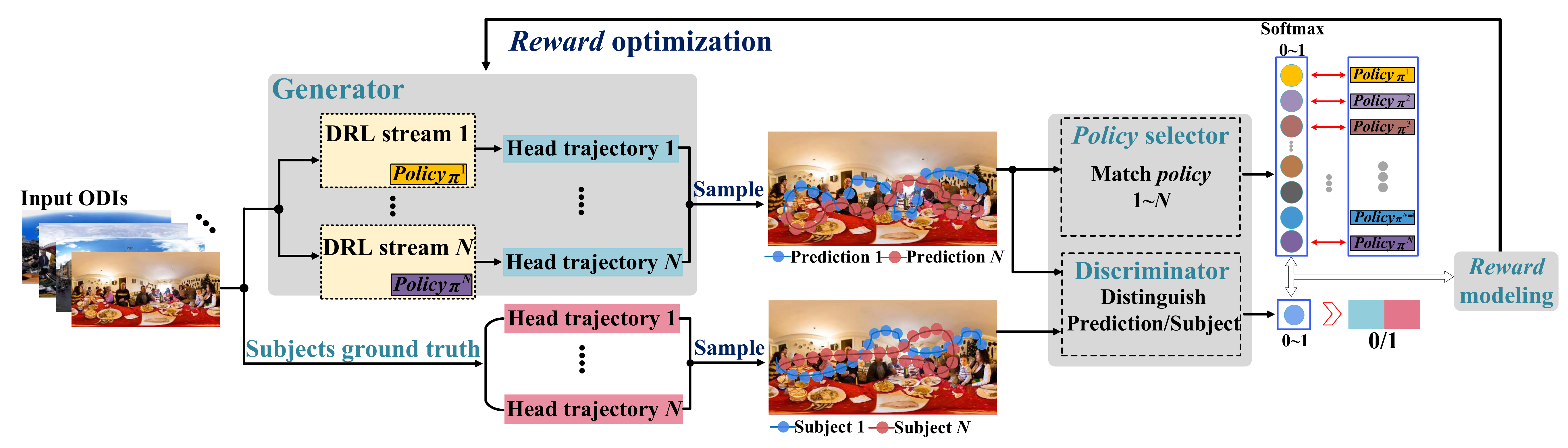}\\
			\centerline{\footnotesize{(a) The training stage of our SalGAIL approach}}\medskip
			\vspace{0.7em}
		\end{minipage}
		\vfill
		\begin{minipage}{1\linewidth}
			\centering
			\includegraphics[width=1.0\linewidth]{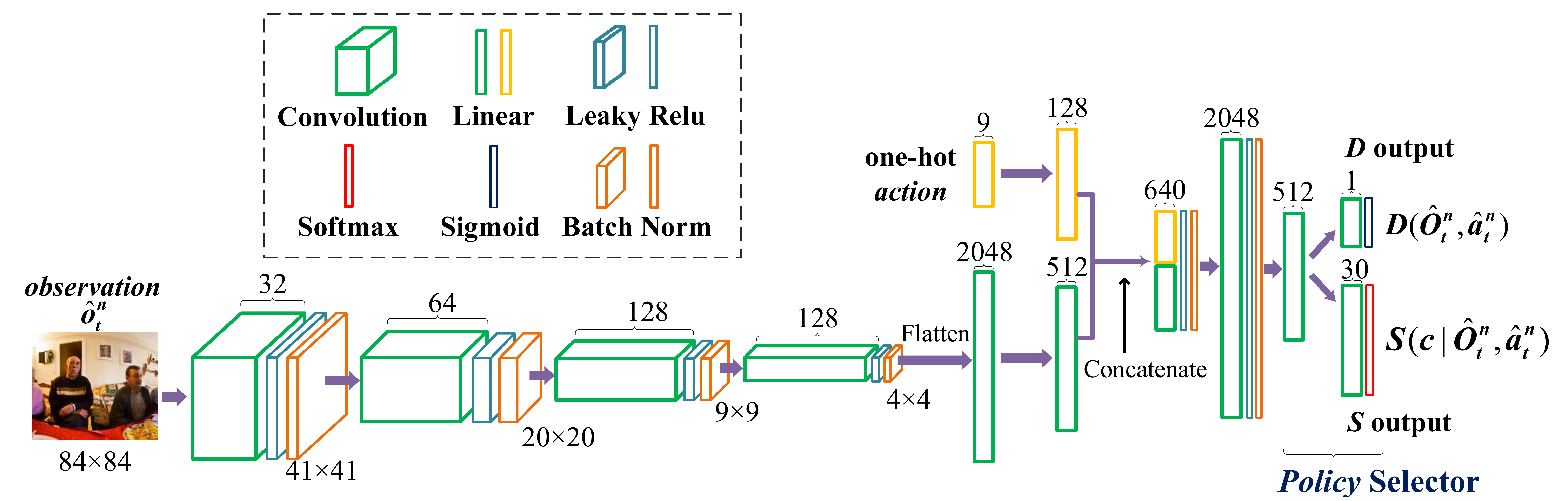}\\
			\centerline{\footnotesize{(b) Structure of CNN in the discriminator and the \textit{policy} selector}}\medskip
		\end{minipage}
		\vfill
		\caption{\footnotesize{(a): The training stage of SalGAIL includes \textit{reward} modeling and optimization. Note that the \textit{reward} \uppercase\expandafter{\romannumeral1} is the output of the discriminator and the \textit{reward} \uppercase\expandafter{\romannumeral2} is the output of the selector. (b): The CNN structure of the discriminator network \textit{D} and posterior approximation network \textit{S}.}}
		\label{fig:train}
		\vspace{-2em}
	\end{figure}

	\item {\textit{Policy} selector.}
	Given a subject, the \textit{policy} selector is used to find a suitable \textit{policy} from one among multiple DRL models, via maximizing the \textit{reward} of imitating the subject's head trajectories.
	This is achieved by adding a latent vector $\mathbf{c}$ into our \textit{policy} function $\pi(\hat{a}\mid\mathbf{\hat{O}}, \mathbf{c})$ as shown in Figure \ref{fig:framework_DRL-CNN}-(b).
	Here, $\pi(\hat{a}\mid\mathbf{\hat{O}}, \mathbf{c})$ is based on InfoGAN \cite{chen2016infogan} and differs from the traditional \textit{policy} function $\pi(\hat{a}\mid\mathbf{\hat{O}})$, which only relies on \textit{observation} $\mathbf{\hat{O}}=\{\hat{\mathbf{O}}_t^n\}_{t=1, n=1}^{T, N}$ and \textit{action} $\mathbf{\hat{a}}=\{\hat{a}_t^n\}_{t=1, n=1}^{T, N}$.
	In our approach, the latent vector $\mathbf{c}$ is represented by a one-hot vector, of which the \textit{n}-th dimension is represented by 1 or 0, corresponding to the \textit{n}-th stream DRL model.
	Guided by $\mathbf{c}$, the \textit{policy} selector can select a specific \textit{policy} from the mixture of \textit{policies} of multiple DRL models.
	Specifically, the \textit{policy} selector acts as a posterior estimation, encouraging the maximum of the mutual information between $\mathbf{c}$ and $\bm{\chi}$.
	In our approach, the posterior estimation can be modeled by the CNN of Figure \ref{fig:train}-(b) as the \textit{policy} selector.
	Note that our discriminator and \textit{policy} selector share the same parameters in first few layers, and the last layer outputs different values for the discriminator and \textit{policy} selector.
	In addition, the mutual information is viewed as one part of the \textit{reward} for the DRL model to be maximized.
	%In our approach, we form \textit{S} as a CNN network which can be seen in Figure \ref{fig:train} (b). Note that our \textit{D} and \textit{S} share the same parameters in first few layers and the last layer outputs different values of \textit{D} and \textit{S}.
	In the following, the details about \textit{reward} modeling are presented.

\end{itemize}
%we form \textit{S} as a CNN network which can be seen in Figure \ref{fig:train} (b).
%	 We use a simplified form $I(\mathbf{c}; \hat{\mathbf{O}}, \hat{a})$ to denote the mutual information. 	
%	Inspired by the information-theoretic generative adversarial network (InfoGAN) \cite{chen2016infogan}
%	
%	Similar to the InfoGAN, we introduce a variational lower bound $L_I(\pi, S)$ of $I(\mathbf{c}; \hat{\mathbf{O}}, \hat{a})$:
%	%\vspace{-0.8em}
%	\begin{align}
%	%\vspace{-1.2em}
%	\label{Mul}
%	\begin{split}
%	L_I(\pi, S) &= E_{\mathbf{c}\sim P(\mathbf{c}), \hat{a}\sim\pi(\hat{a}\mid\hat{\mathbf{O}}, \mathbf{c})}[\log S(\mathbf{c}\mid \hat{\mathbf{O}}, \hat{a})] + H(\mathbf{c})       \\
%	& \leq I(\mathbf{c}; \hat{\mathbf{O}}, \hat{a}).
%	\end{split}
%	\end{align}

%   Consequently, the generative process of the subjects head trajectories can be extended as: $\bm{c}\sim p(\bm{c}), \pi\sim p(\pi\mid \bm{c}), \hat{a}_t\sim \pi(\hat{a}_t\mid\hat{\mathbf{O}}_t, \bm{c}) $,  where $p(\bm{c})$ is the prior distribution of $\bm{c}$ (Which is known before training).
%	
%However, $I(\bm{c}; \hat{\mathbf{O}}, \hat{a})$ is hard to maximize directly as it requires access to the posterior $P(\bm{c}\mid \hat{\mathbf{O}}, \hat{a})$. Therefore,

\textbf{Reward modeling.}
We take into account two components for modeling the \textit{reward}.
The first component is the probability of the predicted head trajectories being those of the subjects, as presented for the discriminator.
The second component is the mutual information between the predicted head trajectories and the latent vector of the \textit{policy} selector.
Specifically, for the \textit{n}-th DRL model of the generator, the $\textit{reward}$ $r_t^n$ at time step $t$ can be defined as follows:
\begin{equation}
\label{reward}
r_t^n = r(\hat{\mathbf{O}}_t^n, \hat{a}_t^n; \bm{\phi}) + \lambda_1 r(\mathbf{c}; \bm{\eta}).
\end{equation}
where $\lambda_1$ ($>0$) is the hyperparameter that controls the trade-off between two components of the \textit{reward}.
In the above equation, $r(\hat{\mathbf{O}}_t^n, \hat{a}_t^n; \bm{\phi})$ is the first component of the \textit{reward}, which is obtained from the output of the discriminator.
Mathematically, it is formulated by
\begin{equation}
\label{r_D}
r(\hat{\mathbf{O}}_t^n, \hat{a}_t^n; \bm{\phi}) = -\log(1- D_{\bm{\phi}}(\hat{\mathbf{O}}_t^n, \hat{a}_t^n)),
\end{equation}
where $D_{\bm{\phi}}$ denotes the CNN of the discriminator with $\bm{\phi}$ as its parameters.
In addition, $r(\mathbf{c}; \bm{\eta})$ is the second component of the \textit{reward} obtained from the \textit{policy} selector, and it is a mutual information as follows,
\begin{equation}
\label{r_S}
r(\mathbf{c}; \bm{\eta})= \sum_{k=1}^{K}p(c_k\mid \mathbf{O}_t^n, a_t^n)\log S_{\bm{\eta}}(c_k\mid \hat{\mathbf{O}}_t^n, \hat{a}_t^n),
\end{equation}
where $S_{\bm{\eta}}$ is the CNN of the \textit{policy} selector with $\bm{\eta}$ as its parameters.
In addition, $c_k$ is the $k$-th element in the $\mathbf{c}$, and  $p(c_k\mid \mathbf{O}_t^n, a_t^n)$ is the probability distribution of the latent vector matching the DRL model to a given subject.
Recall that $K$ is the number of streams in our DRL model.

Next, the accumulated discount \textit{reward} of time step $t$ at each episode can be calculated as follow:
\begin{equation}
\label{V_estimate}
R_t^n = \sum_{b=t}^{B}\gamma^{b-t}r_b^n,
\end{equation}
where $\gamma$ is the discount factor of \textit{Q-learning} \cite{Watkins1992} and $B$ is the step size of one episode.
%	Then, $V^\star(\mathbf{\hat{O}}^n, \hat{a}^n)$=$\{V^\star(\mathbf{\hat{O}}_t^n, \hat{a}_t^n)\}_{t=1}^{T}$ is viewed as the discounted return in one episode, to be used in the \textit{reward} optimization.
%	Additionally, the advantage estimate $A_n(\mathbf{\hat{O}}^n, \hat{a}^n)$ at one episode of each DRL training cycle can be obtained as follows,
%	\begin{equation}
%	\label{A_estimate}
%	A_n(\mathbf{\hat{O}}^n, \hat{a}^n) = V^\star(\mathbf{\hat{O}}^n, \hat{a}^n) - V(\mathbf{\hat{O}}^n, \hat{a}^n).	
%	\end{equation}
%	In the above equation, $V(\mathbf{\hat{O}}^n, \hat{a}^n)$ is the state value that is the average of the accumulated \textit{reward}, and it is obtained from the output of the DRL model.
%	Here, $A_n(\mathbf{\hat{O}}^n, \hat{a}^n)$ measures the deviation between $V^\star(\mathbf{\hat{O}}^n, \hat{a}^n)$ and $V(\mathbf{\hat{O}}^n, \hat{a}^n)$.
Finally,  $r_t^n$ and $R_t^n$ are delivered into the \textit{rollout} and used for updating the parameters of the generator through \textit{reward} optimization. The optimization procedure is discussed in the following.

\textbf{Optimization.}	
Given \eqref{V_estimate}, we optimize the \textit{rewards} of all DRL streams in our SalGAIL approach by maximizing the expectation of the accumulated discount \textit{rewards} at each episode,
\begin{equation}
\label{optim_rew1}
\max\limits_{\bm{\pi}} \mathbb{E}_{\bm{\pi}}[R_t^n] = \max\limits_{\bm{\pi}} \mathbb{E}_{\bm{\pi}}[\sum_{b=t}^{B}\gamma^{b-t}r_b^n],
\end{equation}
such that \textit{policies} $\bm{\pi}$ can be learned in the generator. Here, $\bm{\pi}=\{\pi_{\bm{\omega}}^n\}_{n=1}^N$ denote the \textit{policies} of $N$ DRL streams, which can be learned by updating the CNN parameters of DRL.
Note that a causal entropy regularization term \cite{boularias2011relative} is added in \eqref{optim_rew1} to ensure the exploration in the decision making of DRL. Mathematically, it is written as
\begin{equation}
\label{obj_H}
H(\bm{\pi}) \triangleq \mathbb{E}_{\bm{\pi}}[-\log \bm{\pi}(\hat{\mathbf{a}}\mid \hat{\mathbf{O}}, \mathbf{c})].
\end{equation}		
Recall that $\mathbf{\hat{O}}$ and $\mathbf{\hat{a}}$ are the sets of \textit{observations} $\mathbf{\hat{O}}$=$\{\hat{\mathbf{O}}_t^n\}_{t=1, n=1}^{T, N}$  and \textit{actions} $\mathbf{\hat{a}}$=$\{\hat{a}_t^n\}_{t=1, n=1}^{T, N}$, respectively. Thus, the optimization formulation of \eqref{optim_rew1} is rewritten as
\begin{equation}
\label{final_optim}
\begin{aligned}
& \max\limits_{\bm{\pi}} \mathbb{E}_{\bm{\pi}}[R_t^n] = \max\limits_{\bm{\pi}}
\mathbb{E}_{\bm{\pi}}\left[\sum_{b=t}^{B} \gamma^{b-t} \left(-\log(1- D_{\bm{\phi}}(\hat{\mathbf{O}}_b^n, \hat{a}_b^n)) \right.\right. \\
& +  \left.\lambda_1 \sum_{k=1}^{K}p(c_k\mid \mathbf{O}_b^n, a_b^n)\log S_{\bm{\eta}}  \left. (c_k\mid \hat{\mathbf{O}}_b^n, \hat{a}_b^n) \right) \right]  - \lambda_2 H(\bm{\pi}),
\end{aligned}
\end{equation}
based on \eqref{r_D} and \eqref{r_S}.
Here, $\lambda_2$ ($>0$) is the hyperparameter for balancing the trade-off of the first two terms and the regularization term in \eqref{final_optim}. Consequently, the generator is capable of imitating the head trajectories of subjects through \textit{reward} optimization \eqref{final_optim}, once $D_{\bm{\phi}}(\hat{\mathbf{O}}_b^n, \hat{a}_b^n)$ and $S_{\bm{\eta}} (c_k\mid \hat{\mathbf{O}}_b^n, \hat{a}_b^n)$ have been obtained.
In the following, we present the details about calculation on $D_{\bm{\phi}}(\hat{\mathbf{O}}_b^n, \hat{a}_b^n)$ and $S_{\bm{\eta}} (c_k\mid \hat{\mathbf{O}}_b^n, \hat{a}_b^n)$, which can be achieved through the optimization on the discriminator and the \textit{policy} selector, respectively.

We introduce adversarial training to learn the CNN parameters of the discriminator, such that $D_{\bm{\phi}}(\hat{\mathbf{O}}_b^n, \hat{a}_b^n)$ can be obtained for \eqref{final_optim}.
In adversarial training, the discriminator tries to make the predicted head trajectories distinguishable from the corresponding ground truth trajectories. Mathematically, $D_{\bm{\phi}}(\hat{\mathbf{O}}_b^n, \hat{a}_b^n)$ can be obtained by solving the following optimization formulation:
\begin{equation}
\label{obj_D}	
\max \limits_{\boldsymbol{D}, \boldsymbol{\hat{D}}} \mathbb{E}_{\bm{\pi}_{S}}[\log \boldsymbol{D}] + \mathbb{E}_{\bm{\pi}}[\log(1 - \boldsymbol{\hat{D}})]. 	
\end{equation}	
Recall that $\bm{\pi}$ and $\bm{\pi}_{S}$ are the \textit{policies} of the generator and subjects, respectively. Here, $\boldsymbol{D}=\{D_{\phi}(\mathbf{O}_t^n, a_t^n)\}_{t=1, n=1}^{B, N}$ and $\boldsymbol{\hat{D}}=\{D_{\phi}(\mathbf{\hat{O}}_t^n, \hat{a}_t^n)\}_{t=1, n=1}^{B, N}$ are the sets of outputs from the discriminator in one episode, in which the inputs are the \textit{observation-action} pairs for the ground truth and the prediction.

Next, we learn the CNN parameters of the \textit{policy} selector by optimizing the mutual information between the predicted head trajectories and latent vector to maximum. Then, we can obtain $S_{\bm{\eta}} (c_k\mid \hat{\mathbf{O}}_t^n, \hat{a}_t^n)$ by solving the following optimization formulation:
\begin{equation}
\label{mut_info}
\max \limits_{\mathbf{S}} \mathbb{E}_{\bm{\pi}}[\log \mathbf{S}],
\end{equation}
where $\mathbf{S}$ denotes the set: $\{S_{\bm{\eta}}(c_n\mid \hat{\mathbf{O}}_t^n, \hat{a}_t^n)\}_{t=1, n=1}^{B, N}$.

\begin{table}
	
	\scriptsize	
	\newcommand{\tabincell}[2]{\begin{tabular}{@{}#1@{}}#2\end{tabular}}
	\begin{center}
		\caption{Settings of hyperparameters in our SalGAIL approach.}
		
		\label{tab:salgail_param}
		\begin{tabular}{|c|l|c|}
			\hline \multirow{6}{*}{Generator} & Maximum number of training cycles $H$ & $5\times 10^4$ \\ & The number of episodes $I$ & 42 \\ & The step size of one episode $B$ & 5 \\ & Mini-batch size & 6  \\ & Discount factor $\gamma$ in \eqref{V_estimate}  & 0.99\\ & Initial learning rate & $7\times 10^{-4}$  \\  & The angle of the negative slope in LeakyReLU & 0.01 \\
			
			\hline \multirow{5}{*}{\tabincell{c}{Discriminator \\ \& \textit{Policy} selector}} & Initial learning rate & $2\times 10^{-4}$ \\ & Batch size & 150  \\ & The angle of the negative slope in LeakyReLU & 0.2 \\ & Numerical stability value in BatchNorm & $1\times 10^{-5}$ \\ & Momentum in BatchNorm & 0.1 \\ & Weight decay & $2\times 10^{-3}$ \\
			
			\hline \multirow{2}{*}{Others} & Trade-off hyperparameter for \textit{reward} $\lambda_1$  in \eqref{final_optim}  & $0.7$ \\ & Causal entropy coefficient $\lambda_2$ in \eqref{final_optim} & 0.01  \\
			\hline
		\end{tabular}
	\end{center}
	%	\hline \multirow{3}{*}{AI} & 1. Shallow CNN \cite{Liu16TIP} & 6.189 & -0.316 & -56.41 & -60.23 & -62.62 & -65.04 \\
	%	& 2. Deep CNN \cite{Li17ICME} & 2.249 & -0.105 & -55.44 & -58.23 & -60.12 & -63.79 \\
	%	& 3. ETH-CNN & \textbf{2.247} & \textbf{-0.104} & \textbf{-56.92} & \textbf{-60.38} & \textbf{-63.61} & \textbf{-66.47} \\
	
\end{table}	
After obtaining  $D_{\bm{\phi}}(\hat{\mathbf{O}}_b^n, \hat{a}_b^n)$ and $S_{\bm{\eta}} (c_k\mid \hat{\mathbf{O}}_b^n, \hat{a}_b^n)$, we can solve the optimization problem of \eqref{final_optim}.
This is achieved by updating the parameters of $\bm{\pi}$. As for $\bm{\pi}$, its parameters $\bm{\omega}$=$\{\bm{\omega}_V^i, \bm{\omega}_{\pi}^i \}$ are composed of two parts, where the first part is used to update the state value in the DRL model, and the second part is used to update the \textit{policy} in the DRL model. Therefore, we can obtain the gradient $\bm{\omega}_V^i$:
\begin{equation}
\label{theta_V}	
\Delta_{\bm{\omega}_V^i} = \mathbb{E}_{\bm{\chi}}[\nabla_{\bm{\omega}_V^i}(R_t^n-V_{\bm{\omega}_V^i}(\mathbf{\hat{O}}^n, \hat{a}^n))^2].
\end{equation}
Here, $V_{\bm{\omega}_V^i}(\mathbf{\hat{O}}^n, \hat{a}^n)$ is one part of output of the DRL model. Based on Lemma \ref{lemma1}, the gradient $\bm{\omega}_{\pi}^i$ can be obtained to optimize \eqref{final_optim}, which is as follows:
\begin{equation}
\small
\label{theta_pi}	
\Delta_{\bm{\omega}_\pi^i} = \mathbb{E}_{\bm{\chi}}[\nabla_{\bm{\omega}_{\pi}^i}\log{\pi_{\bm{\omega}_{\pi}^i}^n}(\hat{a}_t^n\mid\hat{\mathbf{O}}_t^n, \mathbf{c}_n)\cdot (\lambda_2 + R_t^n)].
\end{equation}	
\begin{table*}
	\newcommand{\tabincell}[2]{\begin{tabular}{@{}#1@{}}#2\end{tabular}}
	\centering
	\caption{\footnotesize{CC, KL divergence, NSS and AUC results of saliency prediction by our and other approaches over the test set of our AOI dataset. }}\label{biterrorTo1}
	
	\scriptsize{
		\begin{tabular}{|c|c|c|c|p{0.9cm}<{\centering}|p{0.6cm}<{\centering}|p{0.8cm}<{\centering}|p{0.84cm}<{\centering}|c|c|p{0.6cm}<{\centering}|c|c|}
			\hline
			\multirow{2}{*}{Categories} & \multirow{2}{*}{Approaches}
			& \multicolumn{4}{c|}{2D images saliency models} & \multicolumn{6}{c|}{ODIs saliency models} & \multicolumn{1}{c|}{Our model} \\
			\cline{3-13}
			& & \multirow{1}{*}{DVA}  & \multirow{1}{*}{BMS}  & \multirow{1}{*}{SALICON}  & \multirow{1}{*}{MLNet}  & \multirow{1}{*}{BMS360} & \multirow{1}{*}{GBVS360}  & \multirow{1}{*}{Startsev}  & \multirow{1}{*}{Zhu}  & \multirow{1}{*}{DHP}  & \multirow{1}{*}{Battisti}  & \multirow{1}{*}{SalGAIL}  \\
			
			\hline
			
			\multirow{4}{*}{\textbf{\textit{Cityscapes}}}
			& CC & 0.667 & 0.643 & 0.547 & 0.566 & 0.721 & 0.642 & 0.691 & 0.747 & 0.653 & 0.596 & \textbf{0.766} \\
			& KL divergence & 0.578 & 0.549& 0.626 & 0.994 & 0.755 & 0.626 & 0.496 & 0.432 & 0.569 & 0.939 & \textbf{0.345} \\
			& NSS & 1.063 & 0.975 & 0.620 & 0.771 & 1.268 & 1.051 & 0.992 & 1.176 & 0.975 & 0.926 & \textbf{1.447} \\
			& AUC & 0.768 & 0.762 & 0.731 & 0.754 & 0.825 & 0.784 & 0.790 & 0.806 & 0.788 & 0.744 & \textbf{0.841} \\
			
			\hline
			
			\multirow{4}{*}{\textbf{\textit{Indoor Scenes}}}
			& CC & 0.594 & 0.536 & 0.512 & 0.662 & 0.655 & 0.551 & 0.484 & 0.636 & 0.579 & 0.535 & \textbf{0.686} \\
			& KL divergence & 0.662 & 0.642 & 0.683 & 0.689 & 0.546 & 0.627 & 0.665 & 0.552 & 0.649 & 0.835 & \textbf{0.458} \\
			& NSS & 1.115 & 0.789 & 1.109 & 1.258 & 1.209 & 0.917 & 0.768 & 1.230 & 0.954& 0.846 & \textbf{1.449} \\
			& AUC & 0.782 & 0.747 & 0.803 & 0.831 & 0.829 & 0.762 & 0.758 & 0.805 & 0.774 & 0.759 & \textbf{0.848} \\
			
			\hline
			
			\multirow{4}{*}{\textbf{\textit{Human Scenes}}}
			& CC & 0.607 & 0.548 & 0.512 & 0.611 & 0.712 & 0.567& 0.600 & 0.735 & 0.621 & 0.598 & \textbf{0.757} \\
			& KL divergence & 0.545 & 0.530 & 0.586 & 0.727 & 0.639 & 0.513 & 0.472 & \textbf{0.354} & 0.545 & 0.722 & 0.352 \\
			& NSS & 1.209 & 1.072 & 0.858 & 1.249 & 1.430 & 0.965 & 1.259& 1.417 & 1.318 & 1.203 & \textbf{1.603} \\
			& AUC & 0.794 & 0.770 & 0.767 & 0.831 & 0.846 & 0.762 & 0.817 & 0.843 & 0.802 & 0.799 & \textbf{0.859} \\
			
			\hline
			
			\multirow{4}{*}{\textbf{\textit{Natural Landscapes}}}
			& CC & 0.492 & 0.443 & 0.401 & 0.366 & 0.734 & 0.532 & 0.613 & 0.725 & 0.512 & 0.606 & \textbf{0.756} \\
			& KL divergence & 0.625 & 0.779 & 0.827 & 1.381 & 0.401 & 0.664 & 0.572 & 0.478 & 0.688 & 0.929 & \textbf{0.228} \\
			& NSS & 0.875 & 0.809 & 0.438 & 0.489 & 1.502 & 0.894 & 0.931 & 1.205 &0.881 & 0.847 & \textbf{1.725} \\
			& AUC & 0.713 & 0.704 & 0.700 & 0.630 & 0.842 & 0.722 & 0.776 & 0.796 & 0.718 & 0.757 & \textbf{0.863} \\
			
			\hline
			
			\multirow{4}{*}{\textbf{Overall}}
			& CC & 0.590 & 0.557 & 0.511 & 0.589 & 0.714 & 0.590 & 0.595 & 0.727 & 0.591 & 0.589 & $\textbf{0.742}$ \\
			& KL divergence & 0.603 & 0.584 & 0.637 & 0.844 & 0.584 & 0.566 & 0.532 & 0.420 & 0.613 & 0.786 & $\textbf{0.345}$ \\
			& NSS & 1.066 & 0.975 & 0.856 & 1.064 & 1.378 & 0.995 & 1.052 & 1.295 & 1.032 & 1.014 & $\textbf{1.556}$ \\
			& AUC & 0.764 & 0.758 & 0.757 & 0.784 & 0.841 & 0.766 & 0.793 & 0.821 & 0.771 & 0.775 & $\textbf{0.853}$ \\
			
			\hline
		\end{tabular}
		\label{evaluation}
	}
	
\end{table*}
\begin{lemma} \label{lemma1}
	Consider that $R_t^n$ is the accumulated discount \textit{reward}. The gradient $\bm{\omega}_{\pi}^i$ for optimizing \eqref{final_optim} at episode $i$  can be calculated by \eqref{theta_pi}.
	
	\textit{Proof}: See Appendix A.
	%\proof See Appendix \ref{appendix1}.
	
\end{lemma}
To solve the optimization problem of \eqref{obj_D}, the parameters of $D_{\bm{\phi}}$ can be updated through the following gradient:
\begin{equation}
\label{phi}
\Delta_{\bm{\phi}_i} = \mathbb{E}_{\bm{\chi}_S}[\nabla_{\bm{\phi}_i}\log D_{\bm{\phi}_i}(\mathbf{O}, a)] + \mathbb{E}_{\bm{\chi}}[\nabla_{\bm{\phi}_i}\log(1 - D_{\bm{\phi}_i}(\hat{\mathbf{\mathbf{O}}}, \hat{a}))].
\end{equation}
Similarly, the parameters of $S_{\bm{\eta}}$ can be updated through the gradient of objective \eqref{mut_info}:
\begin{equation}
\label{psi}
\Delta_{\bm{\eta}_i} = -\lambda_1\mathbb{E}_{\bm{\chi}}[\nabla_{\bm{\eta}_i}\log S_{\bm{\eta}_i}(\mathbf{c}\mid\mathbf{\hat{O}}, \hat{a})].
\end{equation}

Finally, we can train the CNN parameters of all networks in our SalGAIL approach by solving the optimization problem of \eqref{final_optim}-\eqref{mut_info}. In summary, the training stage of our SalGAIL approach can be seen in Algorithm \ref{alg:GAIL}.

%	Hence, we update $\Delta_{\bm{\omega}_{\pi}^i}$ by policy gradient to increase the objective:
%	\vspace{-0.3em}
%	\begin{equation}
%	\label{theta_pi}	
%	\mathbb{E}_{\bm{\chi}}[\log{\pi_{\bm{\omega}_{\pi}^i}^n}(\hat{a}^n\mid\hat{\mathbf{O}}^n, \mathbf{c}_n)A_n(\hat{\mathbf{O}}^n, \hat{a}^n)]-\lambda_2 H(\pi_{\bm{\omega}_{\pi}^i}^n).
%	\end{equation}

\section{Experimental results}
\subsection{Settings}
\label{settings_zong}
In this section, we validate the effectiveness of the proposed SalGAIL approach.
To this end, each category of  ODIs in our AOI dataset is randomly divided into training and test sets in a ratio of $5:1$.
As a result, there are 500 training ODIs and 100 test ODIs.
Then, we compare the performance of our SaGAIL approach with other state-of-the-art approaches, including DVA \cite{wang2018deep}, BMS \cite{zhang2016bms}, SALICON \cite{huang2015salicon}, MLNet \cite{cornia2016a}, BMS360 \cite{LEBRETON2018GBVS360}, GBVS360 \cite{LEBRETON2018GBVS360}, Startsev \textsl{et al.} \cite{startsev2018360}, Zhu \textsl{et al.} \cite{ZHU2018the}, DHP \cite{xu2018predicting} and Battisti \textsl{et al.} \cite{BATTISTI2018a}.
Among these approaches, DVA, BMS, SALICON, and MLNet are the latest saliency prediction approaches for 2D images.
The remaining approaches are the recent approaches for predicting saliency maps of head fixations on ODIs.
Since only the training models of SALICON, MLNet and DHP are available online, they are retrained over our training set for a fair comparison.

In our SalGAIL approach, the input to the generator at each time step is the predicted viewport at  the last time step,  which has been projected onto a 2D plane and downsampled to $84 \times 84$.
The number of streams $N$ was set to 30 in our SalGAIL approach, the same as the subject number.
When training the generator of our SalGAIL approach, the hyperparameters were tuned to optimize the accumulated discount \textit{rewards} over the training set.
The values of these hyperparameters can be found in Table \ref{tab:salgail_param}.
Table \ref{tab:salgail_param} further tabulates the key hyperparameter settings of the discriminator and the \textit{policy} selector, also tuned over the training ODIs.
Then, the RMSprop optimizer \cite{tieleman2012lecture} and the Adam optimizer \cite{kingma2014adam} were used to update the parameters of the generator and the discriminator/\textit{policy} selector, respectively.
All experiments were conducted on a computer with an Intel(R) Core(TM) i7-6700K CPU@4.0 GHz, 32 GB of RAM and a single Nvidia GeForce GTX 1080Ti GPU.

%	
%	\begin{table*}[!tb]
%		\centering
%		\caption{ CC, KL, NSS and AUC results of saliency prediction by our and other approaches over all images in test set. The best result is highlighted in bold.}
%		%\vspace{-.5em}
%		\vspace{3pt}
%		\resizebox{1\linewidth}{!}{
%			\begin{tabular}{ccccccccccccc}
%				\toprule
%				& Ours & DVA & SalGAN& SALICON & BMS  & MLNet & BMS360 & GBVS360 & TUM & SJTU & NWPU & Roma \\
%				\midrule
%				\multirow{1}[2]{*}{CC} & \textbf{0.716} & XXX  & 0.693 & 0.511 & 0.566 & 0.495 & 0.693 & 0.589 & 0.691  & 0.698  & 0.703  & 0.675 \\
%				\multirow{1}[2]{*}{KL}  & \textbf{1.185} & XXX & 1.324 & 2.484 & 2.346 & 2.622 & 1.307 & 1.856 & 1.331  & 1.289  & 1.245  & 1.552 \\
%				\multirow{1}[2]{*}{NSS}  & \textbf{1.679} & XXX & 1.483 & 0.856 & 0.974 & 0.687 & 1.125 & 0.991& 1.186  & 1.496  & 1.525 & 1.045   \\
%				\multirow{1}[2]{*}{AUC}  & \textbf{0.889} & XXX & 0.875 & 0.824 & 0.842 & 0.819 & 0.877 & 0.850 & 0.873 & 0.877 & 0.885 & 0.871 \\	
%				\bottomrule
%			\end{tabular}
%		}
%		\label{matric_table}
%		\vspace{-1em}
%	\end{table*}

\subsection{Performance evaluation on SalGAIL}
%Here, we report our experiment results about the performance evaluation on SalGAIL from 3 ways, which is objective, subjective and generalization  performance evaluation.
\textbf{Objective evaluation}.
In our experiments, we objectively evaluate the accuracy of saliency prediction of head fixations in terms of four metrics: CC, KL divergence, normalized scanpath saliency (NSS) and the area under the receiver operating characteristic curve (AUC).
The larger values of CC, NSS or AUC indicate higher accuracy in saliency prediction, while a smaller KL divergence means better performance of saliency prediction.
Table \ref{evaluation} tabulates the results of CC, KL divergence, NSS and AUC for our own and 10 other approaches over each category of ODIs and all ODIs.
We can see from this table that our SalGAIL approach performs much better than other approaches in terms of different metrics.
In particular, our SalGAIL approach has a 0.015 increase in CC, 0.075 reduction in KL divergence, 0.178 increase in NSS and 0.012 increase in AUC, over Zhu \textsl{et al.}  and BMS360 which perform the best among all compared approaches.
In addition, our approach is also superior to other approaches in four metrics for categories of \textit{cityscapes, indoor scenes, human scenes} and \textit{textnatural landscapes}.
For \textit{human scenes}, our approach achieves the best performance in terms of CC, NSS and AUC, while its KL divergence result is slightly worse than those of Zhu \textsl{et al.}.
Generally, our SalGAIL approach is effective in saliency prediction of head fixations on ODIs and is superior to other state-of-the-art approaches.
\begin{figure*}
	\label{subjective fig}
	\centering
	\includegraphics[width=1\linewidth]{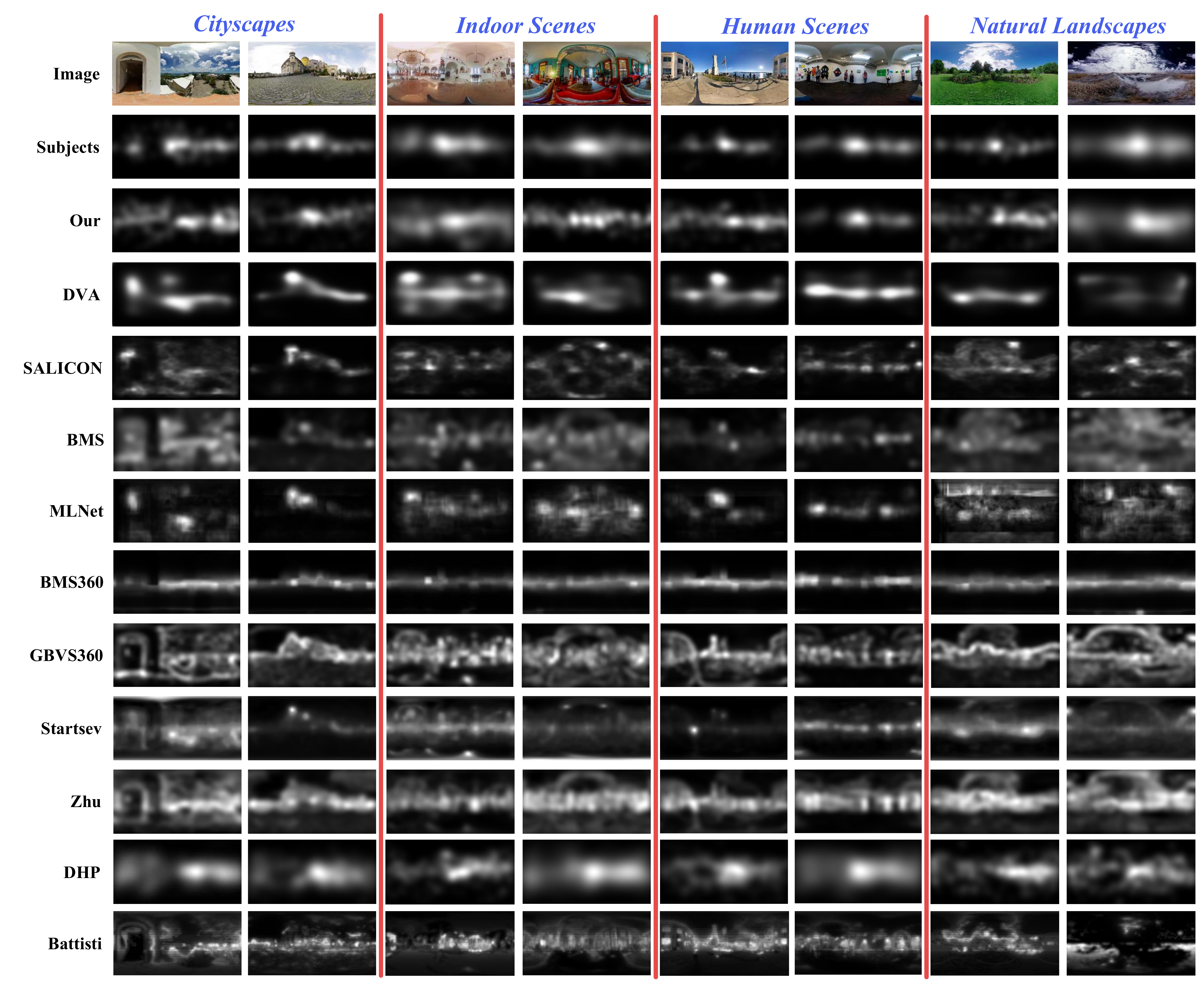}
	\caption{Examples of saliency maps of head fixations, predicted by our SalGAIL and other approaches.}
	\label{fig:matric1}
\end{figure*}

\textbf{Subjective evaluation}.
Next, we compare the subjective results of saliency prediction on ODIs. For each category of ODIs, 2 test ODIs are randomly selected from our AOI dataset.
Figure \ref{fig:matric1} visualizes the saliency maps of the selected ODIs generated by our own and 10 other approaches.
As seen in this figure, the saliency maps of our SalGAIL approach are much closer to those of the ground truth, when compared with other approaches.
This result implies that our SalGAIL approach performs well in the subjective results of saliency prediction.
In summary, both objective and subjective results show that our SalGAIL approach outperforms other state-of-the-art approaches for predicting saliency of head fixations on ODIs.

\subsection{Generalization ability test}
\textbf{Evaluation over the Salient360 dataset}. Now, we assess the generalization ability of our approach by testing over the Salient360 dataset \cite{Rai2017a}. Here, the SalGAIL model, learned from the training set of our AOI dataset, is directly used to predict the saliency maps of head fixations on all ODIs of the Salient360 dataset. We also compare the performance of our SalGAIL approach with 10 other approaches in terms of CC, KL divergence, NSS and AUC. The average results are reported in Table \ref{gener}. As shown in this table, our SalGAIL approach again outperforms all other approaches. Specifically, there is at least a 0.021 increase in CC, 0.095 reduction in KL divergence, 0.085 increase in NSS, and 0.013 increase in AUC.
This result demonstrates the high generalization ability of our SalGAIL approach.

\textbf{Evaluation over the VR dataset}. We further test the performance of our SalGAIL approach over the VR dataset \cite{Sitzmann2018saliency}.  In the VR dataset, two groups of head fixations are collected, for viewing the same ODIs in the \textit{standing} and \textit{seated} conditions, respectively. Note that the head fixations are obtained in the \textit{seated} condition for both our AOI dataset and the Salient360 dataset. Consequently, the generalization ability of our SalGAIL approach can be evaluated for different viewing conditions. The average results of the CC, KL divergence, NSS and AUC are also reported in Table \ref{gener}. We can see from this table that our SalGAIL approach also performs better than all other approaches, for both \textit{standing} and \textit{seated} conditions.
In summary, our SalGAIL approach has higher generalization ability on different datasets and in different viewing conditions.

\begin{table*}[!tb]
	\centering
	\caption{Mean (standard deviation) values for saliency prediction accuracy of our and other approaches over Salient360 and Saliency in VR datasets.}
	%\vspace{-.5em}
	%	\resizebox{1\linewidth}{!}{
	\begin{tabular}{|c|ccccccccccc|}
		\hline
		\multirow{2}{*}{} & \multicolumn{11}{c|}{\textbf{Salient360 (\textit{Seated} condition)}} \\
		\hline
		& Ours & DVA & BMS & SALICON   & MLNet & BMS360 & GBVS360 & Startsev & Zhu & DHP & Battisti \\
		\cline{2-12}
		\multirow{1}[2]{*}{CC} & \textbf{0.757} & 0.534 & 0.502 & 0.467 & 0.429 & 0.736 & 0.502 & 0.612 & 0.678 & 0.565  & 0.563 \\
		\multirow{1}[2]{*}{KL divergence} & \textbf{0.366} & 0.727 & 0.597 & 0.764 & 1.367 & 0.647 & 0.642 & 0.555 & 0.461 & 0.738 & 0.742\\
		\multirow{1}[2]{*}{NSS} & \textbf{0.893} & 0.643 & 0.665 & 0.387 & 0.462 & 0.808 & 0.607 & 0.479 & 0.720 & 0.658  & 0.652 \\
		\multirow{1}[2]{*}{AUC} & \textbf{0.708} &0.661  & 0.655 & 0.633 & 0.638 & 0.695 & 0.642 & 0.637 & 0.691 & 0.664 & 0.665 \\
		\hline
		\multirow{2}{*}{} & \multicolumn{11}{c|}{\textbf{Saliency in VR (\textit{Standing} condition)}} \\
		\hline
		& Ours & DVA & BMS & SALICON & MLNet & BMS360 & GBVS360 & Startsev & Zhu & DHP & Battisti \\
		\cline{2-12}
		\multirow{1}[2]{*}{CC} & \textbf{0.641} & 0.502 & 0.475 & 0.444 & 0.458 & 0.627 & 0.286 & 0.441  & 0.544  &  0.475 & 0.523 \\
		\multirow{1}[2]{*}{KL divergence} & \textbf{0.425} &0.618  & 0.644 & 0.686 & 1.169 & 0.495 & 0.989 & 0.815  & 0.586  & 0.764  & 0.626 \\
		\multirow{1}[2]{*}{NSS} & \textbf{1.467} & 1.086 & 1.072 & 1.061 & 1.052 & 1.413 & 0.624& 1.059  & 1.198  & 1.124 & 1.150 \\
		\multirow{1}[2]{*}{AUC} & \textbf{0.747} & 0.705 & 0.694 & 0.678 & 0.665 & 0.738 & 0.608 & 0.665 & 0.711  &0.689 & 0.702\\
		\hline		
		\multirow{2}{*}{} & \multicolumn{11}{c|}{\textbf{Saliency in VR (\textit{Seated} condition)}} \\
		\hline
		& Ours & DVA & BMS & SALICON & MLNet & BMS360 & GBVS360 & Startsev & Zhu & DHP & Battisti \\
		\cline{2-12}
		\multirow{1}[2]{*}{CC} & \textbf{0.603} &0.376  & 0.356 & 0.295 & 0.344 & 0.518& 0.370 & 0.261  & 0.586  & 0.421 &0.503 \\
		\multirow{1}[2]{*}{KL divergence} & \textbf{0.454} &1.157  & 1.102 & 0.946 & 1.489 & 1.861 & 0.905 & 1.286  & 0.492  & 0.923 & 1.110 \\
		\multirow{1}[2]{*}{NSS} & \textbf{1.042} &0.635 & 0.576 & 0.510 & 0.567 & 0.970 & 0.624& 0.571  & 0.977  & 0.745 & 0.815  \\
		\multirow{1}[2]{*}{AUC} & \textbf{0.772} & 0.671 & 0.665 & 0.669 & 0.665 & 0.760 & 0.669 & 0.641 & 0.742 & 0.695 & 0.718 \\
		\hline	
		
	\end{tabular}
	%	}
	\label{gener}
\end{table*}

\subsection{Ablation analysis}

\begin{figure}
	\centering
	\includegraphics[width=0.95\linewidth]{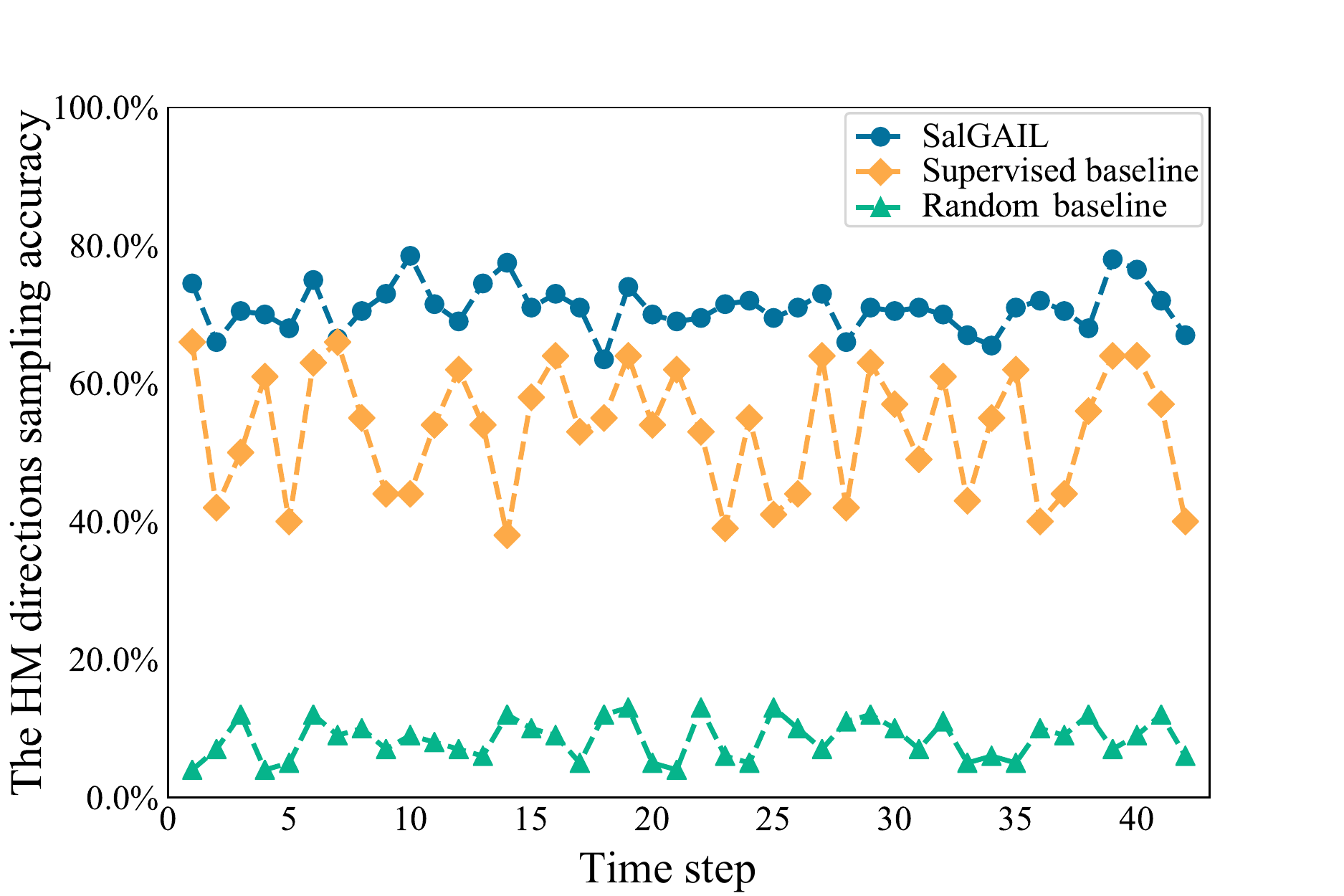}
	\caption{Accuracy of predicting HM directions alongside time steps.}
	\label{acr}		
\end{figure}

\textbf{Ablation on DRL.}
We evaluate the effectiveness of DRL applied in our SalGAIL approach, via replacing it by the supervised learning baseline. Specifically, the supervised baseline acts as a classifier, which is modeled by the CNN, to predict the 8 discrete HM directions or \textit{stay} in the \textit{action} space. For a fair comparison, the CNN structure of the supervised baseline is the same as that of the DRL model in our SalGAIL approach. Meanwhile, the input to the supervised baseline is the viewport extracted from the ODI, which is also the same as the input to SalGAIL. Additionally, the output of the supervised baseline is the probability distribution over 8 discrete HM directions and \textit{stay} in the \textit{action} space. Similar to our SalGAIL approach, the supervised baseline also runs 30 streams of classifiers to predict the head fixations of 30 subjects.
In the training stage, for one stream, the supervised baseline learns one classifier model from the ground truth data of the corresponding subject.
In the test stage, the baseline selects an \textit{action} for one stream at each time step, based on the corresponding trained classifier.
In addition to the supervised baseline, we also compare SalGAIL with the random baseline, in which the action of HM directions and \textit{stay} are randomly generated.
Figure \ref{acr} shows the accuracy of predicting HM directions along with time steps, for our SalGAIL approach, the supervised baseline and the random baseline.

\begin{table}[!tb]
	\centering
	\caption{Comparison of saliency prediction results among our SalGAIL approach, the supervised baseline and the random baseline.}
	%\vspace{-.5em}
	\vspace{3pt}
	%	\resizebox{1\linewidth}{!}{
	\begin{tabular}{|ccp{1.8cm}<{\centering}p{1.8cm}<{\centering}|}
		\hline
		& \multirow{1}{*}{SalGAIL} & Supervised & Random\\
		\hline
		
		CC & \textbf{0.742} & 0.524 &  0.235 \\
		
		KL divergence & \textbf{0.345} & 0.754 & 1.867 \\
		
		NSS & \textbf{1.556} & 0.612 & 0.325 \\
		
		AUC & \textbf{0.853} & 0.593 & 0.387  \\
		\hline
	\end{tabular}
	%	}
	\label{rl_dl_metric}
	\vspace{-1em}
\end{table}
We can see that the prediction accuracy dramatically decreases when replacing DRL by the supervised baseline and random baseline.
Moreover, the predicted head trajectories can be obtained upon the predicted \textit{actions} from all streams, and then the saliency map is generated by convoluting all head fixations which are sampled from the predicted head trajectories.
Table \ref{rl_dl_metric} tabulates the CC, KL divergence, NSS and AUC values of SalGAIL, the supervised baseline and the random baseline.
As shown in Table \ref{rl_dl_metric}, the proposed SalGAIL approach performs much better than both the supervised and random baselines.
This result validates the effectiveness of DRL applied in our SalGAIL.

\begin{figure}
	\begin{minipage}{0.495\linewidth}
		\centering
		\includegraphics[width=45mm]{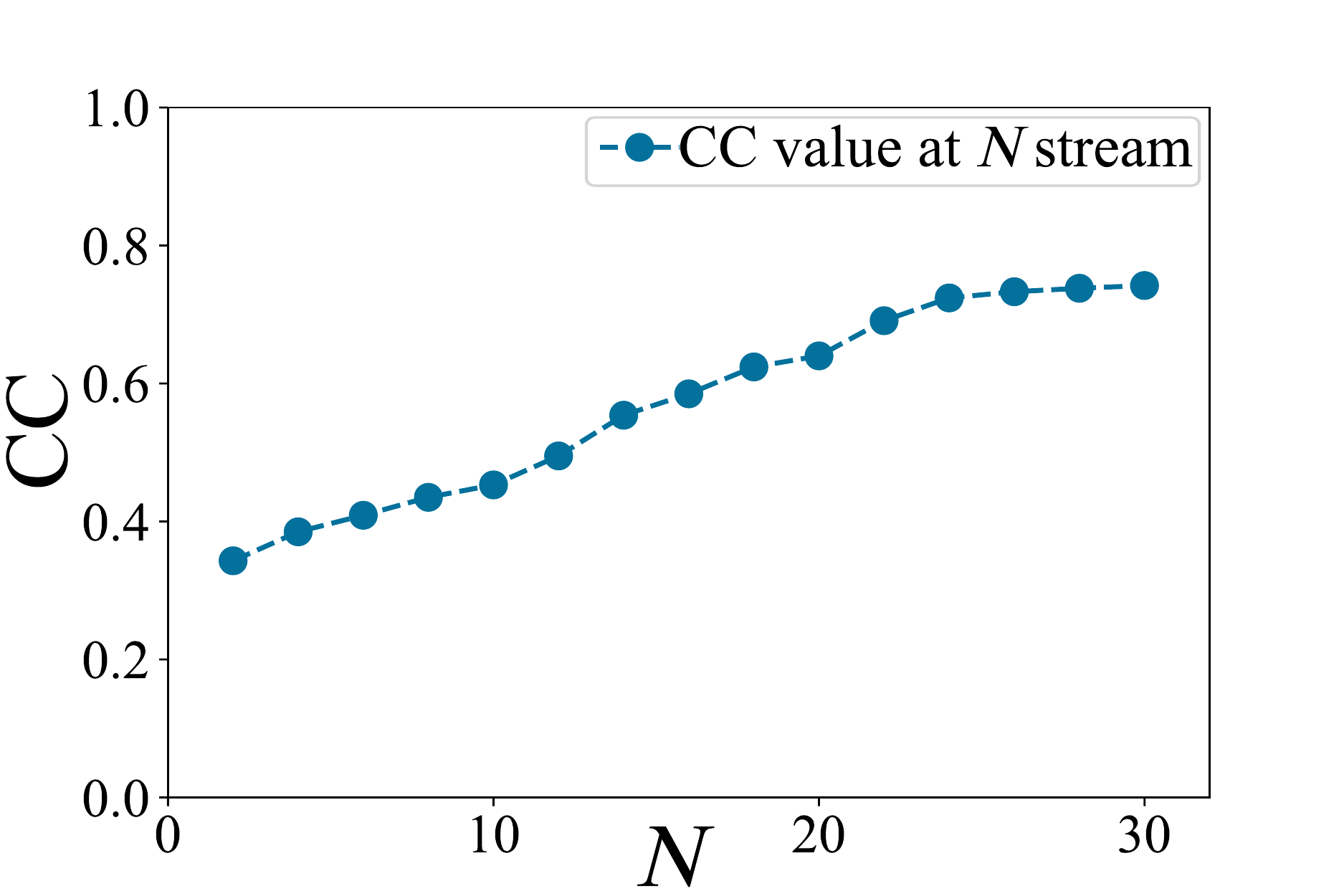}\\
		\vspace{0.9em}
	\end{minipage}
	\hfill
	\begin{minipage}{0.495\linewidth}
		\centering
		\includegraphics[width=43mm]{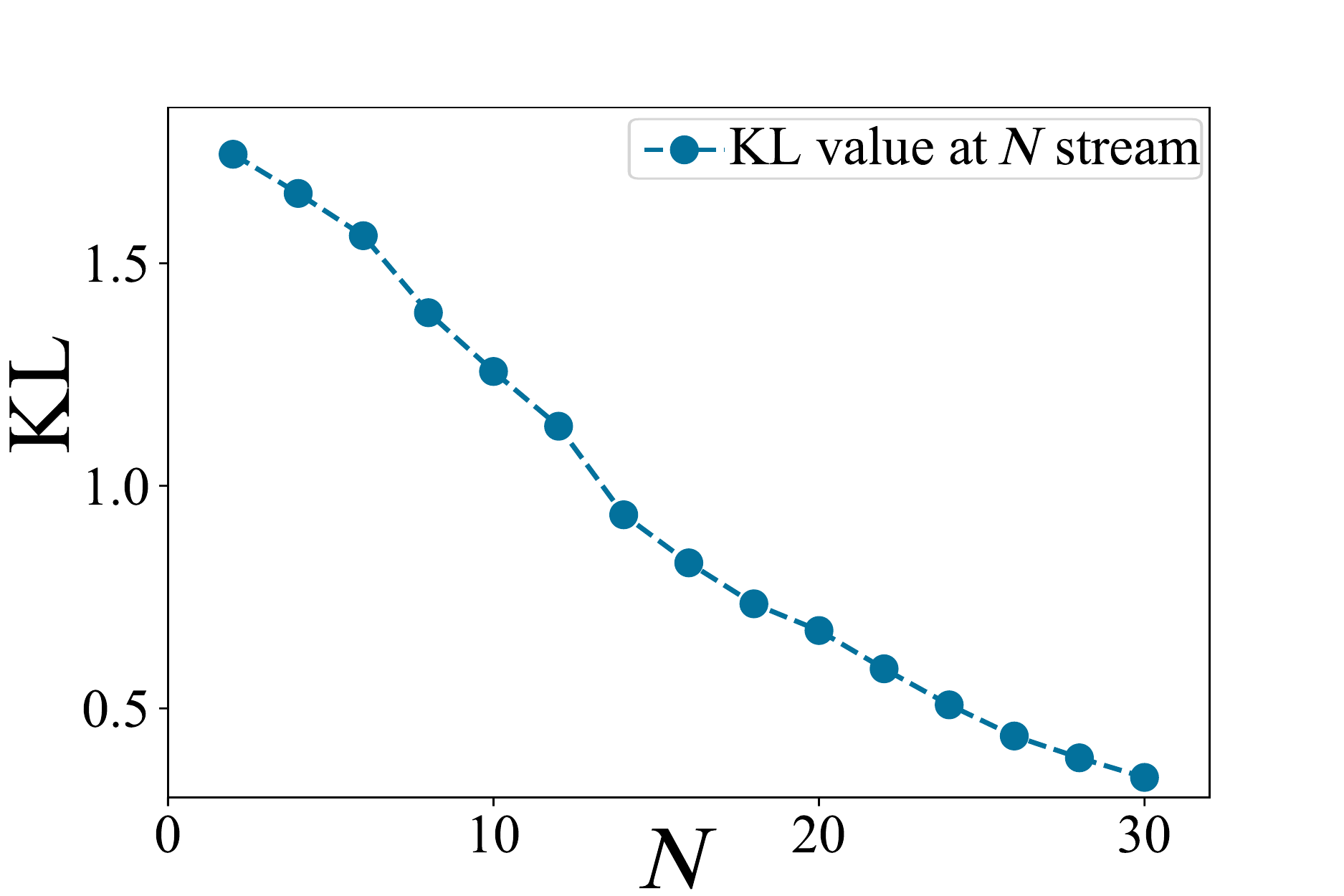}\\
		\vspace{0.7em}
	\end{minipage}
	\vfill
	\begin{minipage}{0.495\linewidth}
		\centering
		\includegraphics[width=45mm]{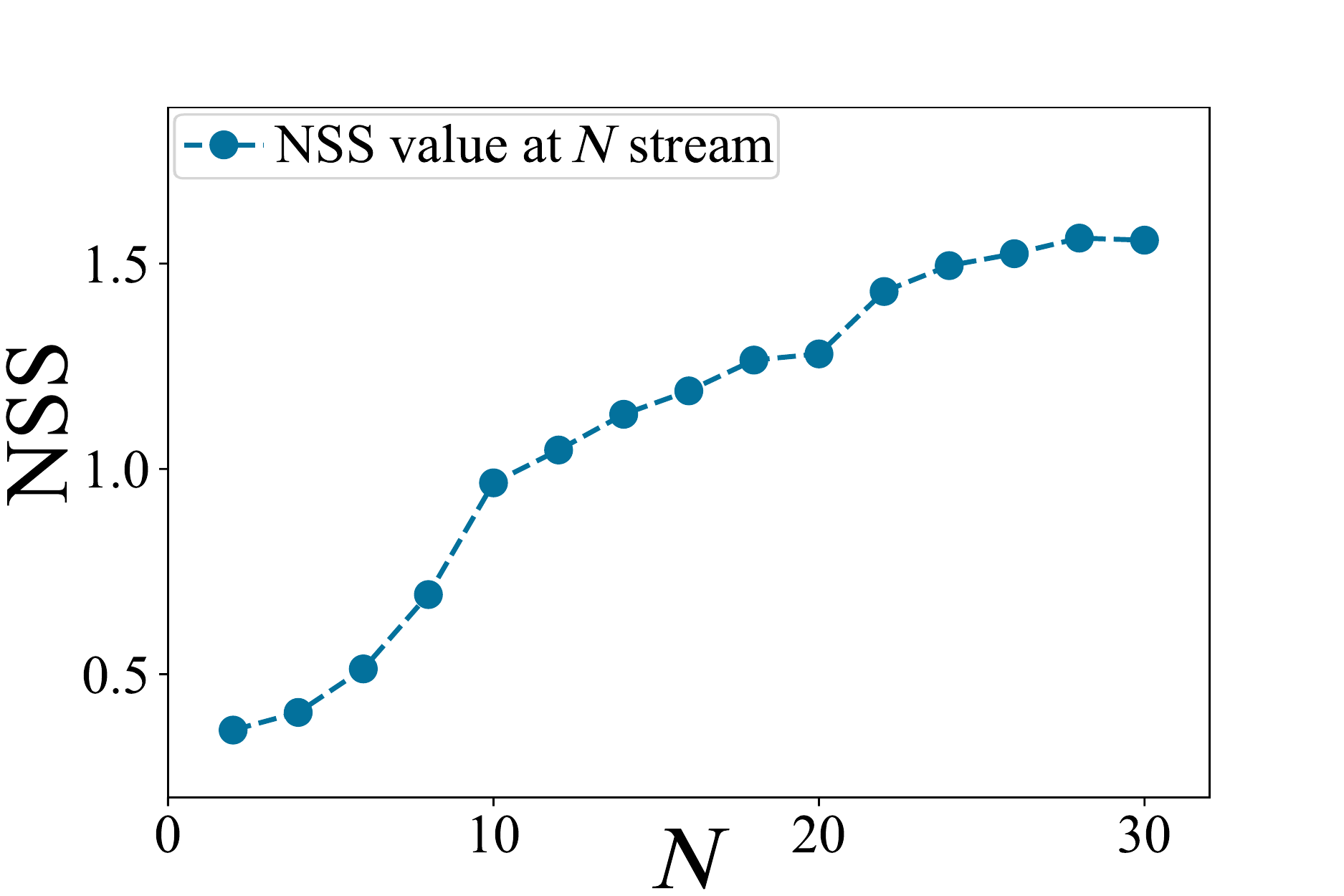}\\
		\vspace{0.7em}
	\end{minipage}
	\hfill
	\begin{minipage}{0.495\linewidth}
		\centering
		\includegraphics[width=45mm]{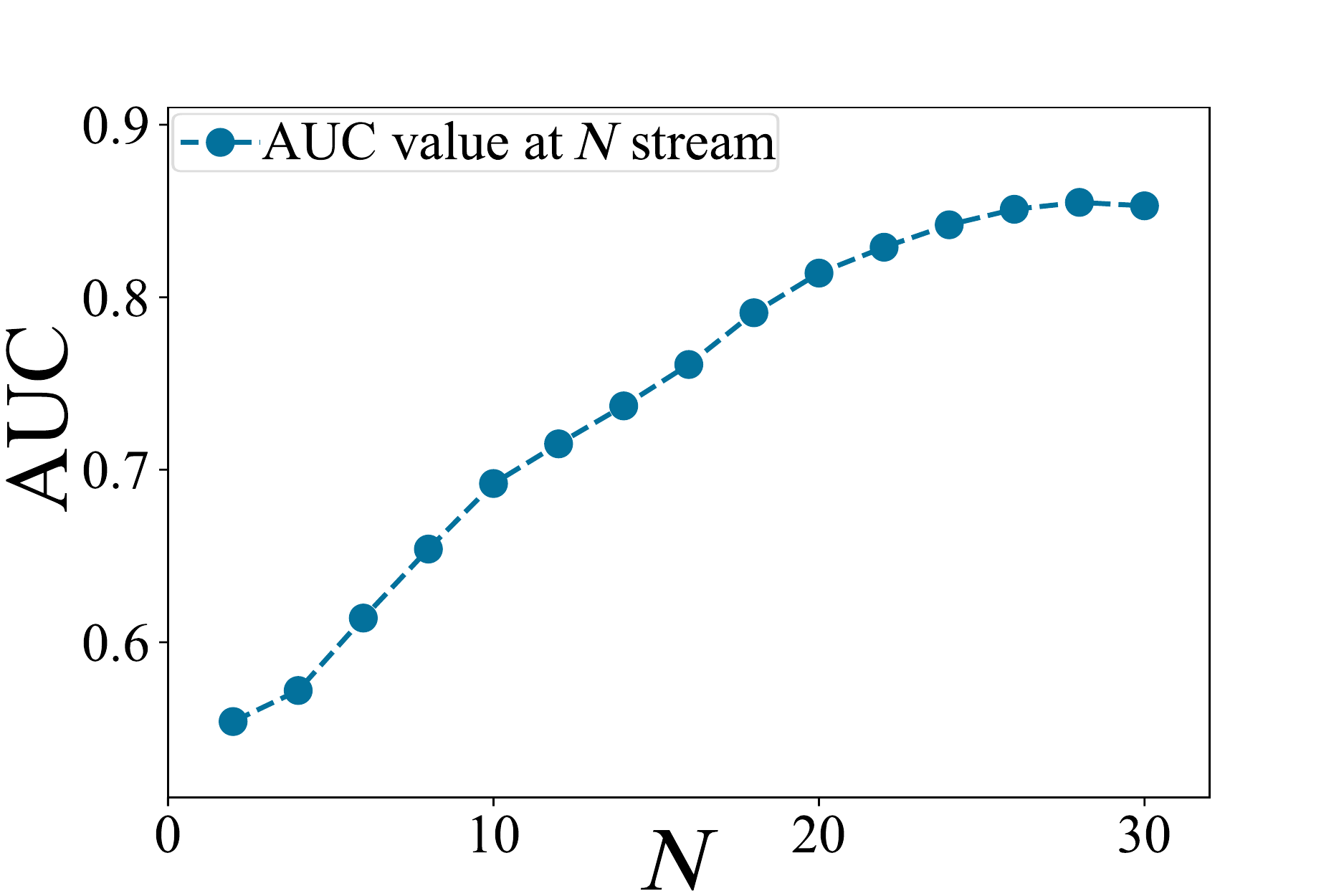}\\
		\vspace{0.7em}
	\end{minipage}		
	\caption{Performance of SalGAIL at different numbers of DRL streams.}
	\label{N STREAMS}
	\vspace{-1em}
\end{figure}

\textbf{Ablation on the stream number of DRL models.}
In our SalGAIL approach, the multi-stream DRL model is used to imitate the head fixations of different subjects.
Here, we conduct the ablation experiments to investigate the influence of DRL stream numbers on the performance of our SalGAIL approach.
The results are plotted in Figure \ref{N STREAMS}.
As shown in this figure, the values of NSS, CC and AUC grow and the KL divergence decreases, along with the increased stream number.
Additionally, all four metrics converge when the stream number of DRL approaches 30.
This result implies the necessity of the multi-stream DRL and the reasonableness of setting the stream number to 30.

\begin{figure}
	\begin{minipage}{0.495\linewidth}
		\centering
		\includegraphics[width=49mm]{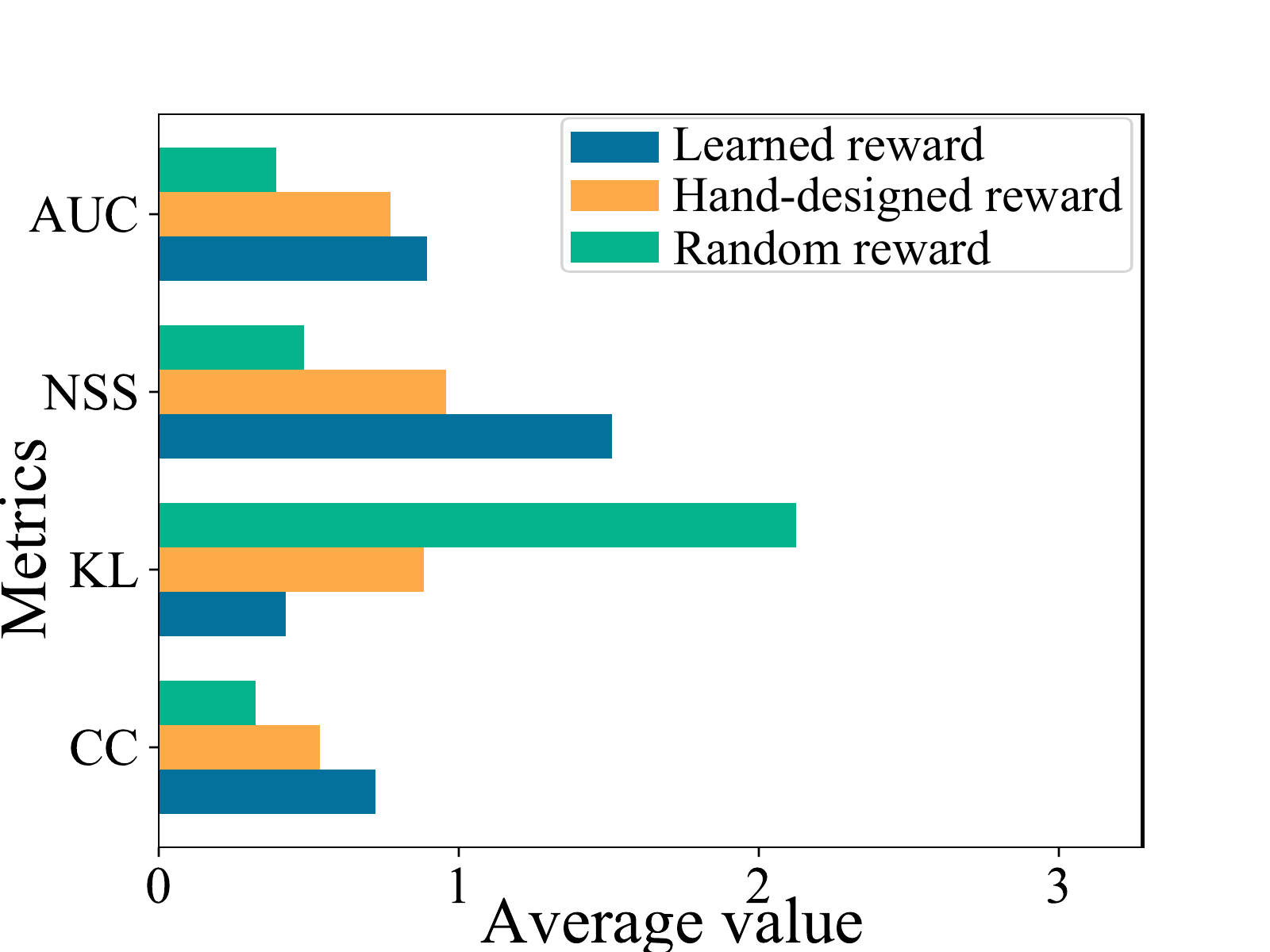}\\
		\centerline{\footnotesize{(a) \textit{Cityscapes}}}\medskip
		\vspace{0.4em}
	\end{minipage}
	\hfill
	\begin{minipage}{0.495\linewidth}
		\centering
		\includegraphics[width=49mm]{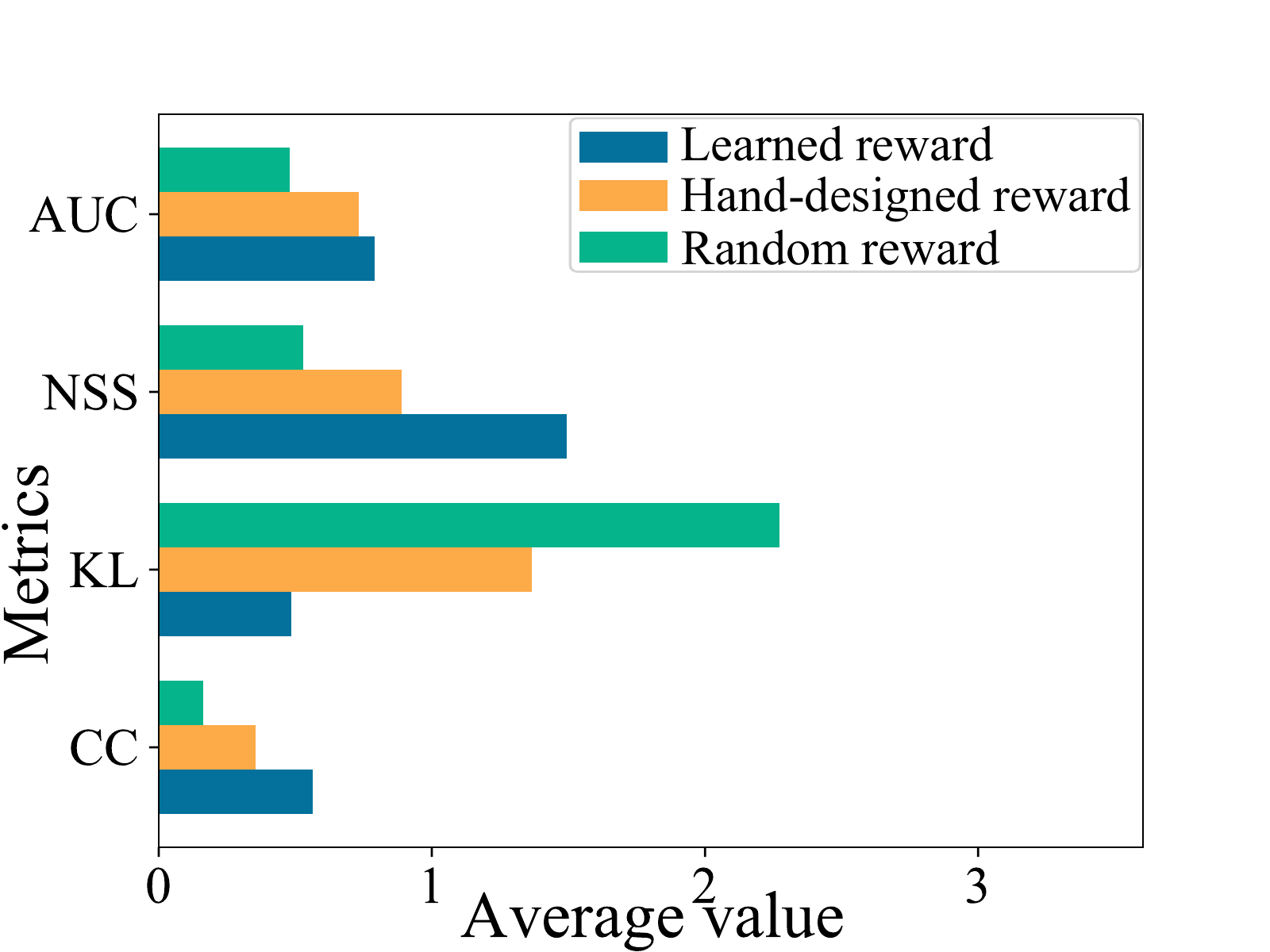}\\
		\centerline{\footnotesize{(B) \textit{Indoor Scenes}}}\medskip
		\vspace{0.4em}
	\end{minipage}
	\vfill
	\begin{minipage}{0.495\linewidth}
		\centering
		\includegraphics[width=49mm]{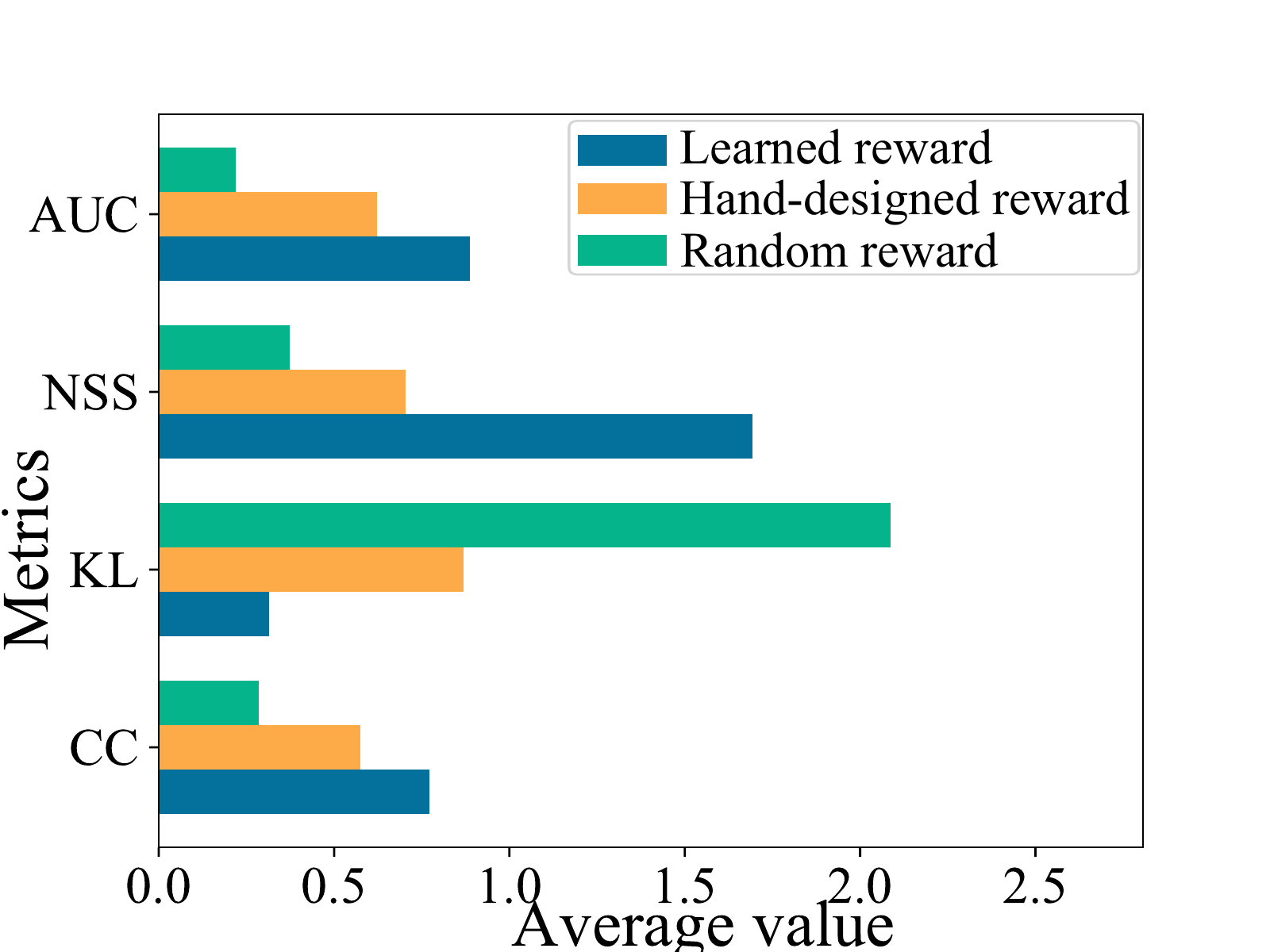}\\
		\centerline{\footnotesize{(c) \textit{Human Scenes}}}\medskip
		\vspace{0.4em}
	\end{minipage}
	\hfill
	\begin{minipage}{0.495\linewidth}
		\centering
		\includegraphics[width=49mm]{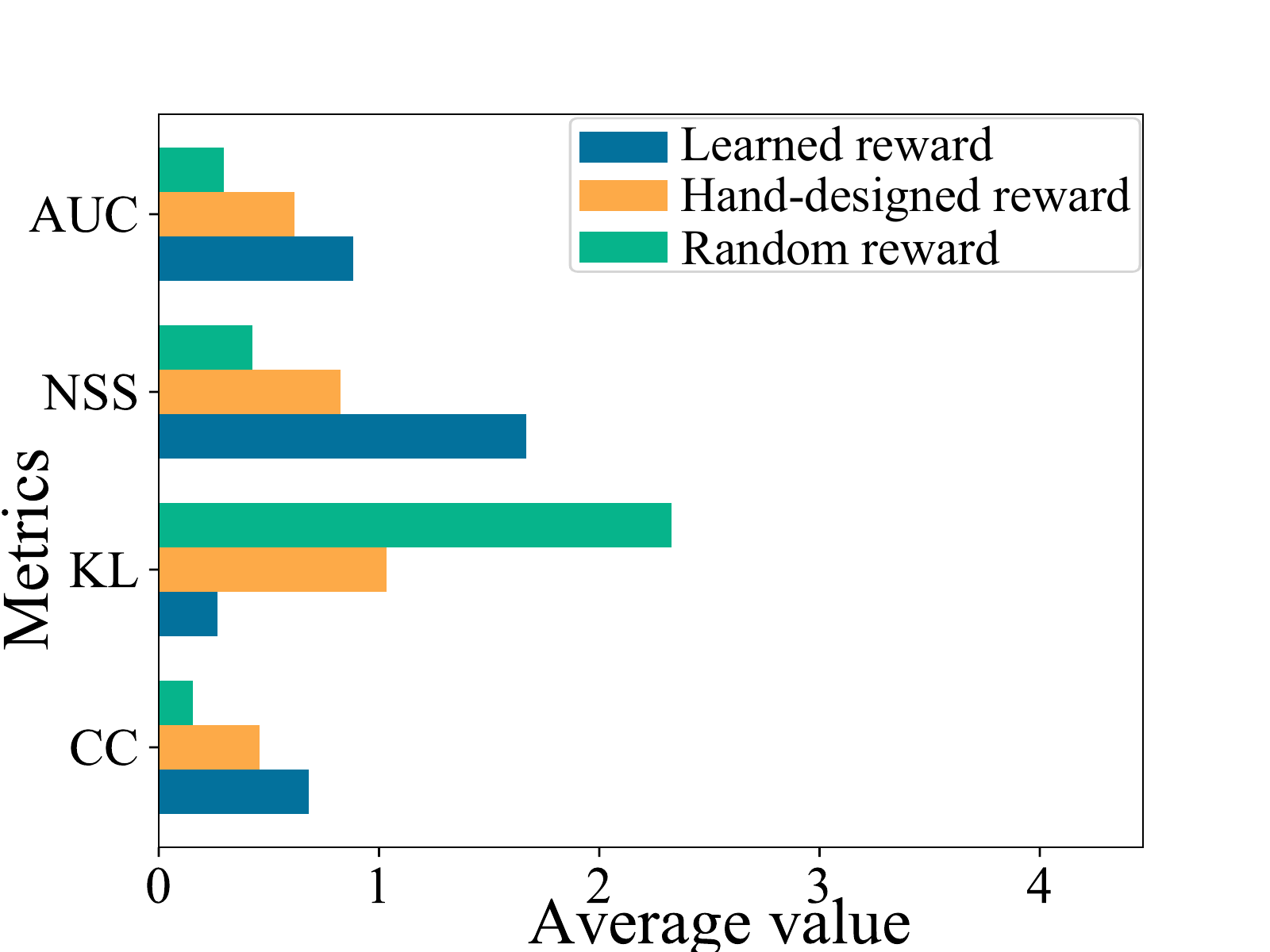}\\
		\centerline{\footnotesize{(d) \textit{Nature landscapes}}}\medskip
		\vspace{0.4em}
	\end{minipage}	
	
	\caption{Performance evaluation over the DRL model that uses learned \textit{reward}, hand-designed \textit{reward} and random \textit{reward} on four categories in AOI.}
	\label{metric_four_category}
	\vspace{-1em}
\end{figure}

\textbf{Ablation on the learned \textit{Reward}.}
Here, we evaluate the effectiveness of the \textit{reward} learned by the GAIL algorithm of our approach. To this end, we replace the learned \textit{reward} by a hand-designed \textit{reward} \cite{xu2018predicting}.
Then, we obtain the performance of our approach with the learned and hand-designed \textit{rewards}.
In addition, we perform a comparison with a random \textit{reward} approach, which randomly samples \textit{actions} at each time step during the whole training process.
The comparison results are shown in Figure \ref{metric_four_category}.
We can see from this figure that the learned \textit{reward} makes our approach perform significantly better than both hand-designed and random \textit{rewards}.
Therefore, the proposed GAIL algorithm of our approach is effective in modeling the \textit{reward} of DRL for saliency prediction on ODIs.

\textbf{Ablation on the FCB.}
Finally, we ablate the FCB in our SalGAIL approach to investigate its effect on saliency prediction of ODIs.
Specifically, we remove the FCB map in \eqref{cb1}, such that the results of our approach without the FCB can be obtained.
Consequently, after removing the FCB, there is a 0.027 reduction in CC, 0.045 increase in KL divergence, 0.085 reduction in NSS and 0.018 reduction in AUC.
Thus, it is necessary to embed the FCB in our SalGAIL approach.

\section{Conclusion}	
In this paper, we have proposed the SalGAIL approach for predicting the saliency maps of head fixations on ODIs.
First, we established the AOI dataset, which is composed of both head fixations and eye fixations of 30 subjects on 600 ODIs.
To the best of our knowledge, AOI is the largest dataset for attention modeling on ODIs.
Second, we mined the AOI dataset and achieved several findings regarding the head fixations of subjects when viewing ODIs.
Third, inspired by these findings, we proposed a multis-tream DRL model in our SalGAIL approach for saliency prediction on ODIs, in which the \textit{reward} is learned by imitating head trajectories of human through the proposed GAIL algorithm.
In the multi-DRL model, each DRL stream yields the HM trajectory of one subject, and then head fixations can be sampled from the yielded HM trajectories of all DRL streams. The experiment also validates the high generation ability of our SalGAIL approach across different datasets.
Finally, a processing technique was presented to convolute the predicted head fixations of an input ODI with a Gaussian kernel, such that the saliency map of the ODI can be generated as the output of our SalGAIL approach.
The extensive experiments showed that our SalGAIL approach is superior to 10 state-of-the-art approaches in predicting the saliency maps of head fixations on ODIs.

There are two promising research directions of future work. (1) The proposed SalGAIL in this paper mainly focuses on saliency prediction of head fixations on ODIs. Saliency prediction of eye fixations remains to be developed for ODIs. This is an important area for future work.
(2) Our SalGAIL approach may be used to remove visual redundancy for some ODI processing tasks, e.g., ODI quality assessment and compression. This is another interesting application for future work.	

{\small
\bibliographystyle{ieee}
\bibliography{egbib}

\begin{thebibliography}{10}\itemsep=-1pt

\bibitem{Reina2017SaltiNet}
M.~{Assens}, X.~{Giro-i-Nieto}, K.~{McGuinness}, and N.~E. {OConnor}.
\newblock Saltinet: Scan-path prediction on 360 degree images using saliency
  volumes.
\newblock In {\em 2017 IEEE International Conference on Computer Vision
  Workshops (ICCVW)}, pages 2331--2338, Oct 2017.

\bibitem{Auer2002}
P.~Auer, N.~Cesa-Bianchi, and P.~Fischer.
\newblock Finite-time analysis of the multiarmed bandit problem.
\newblock {\em Machine Learning}, 47(2):235--256, May 2002.

\bibitem{BATTISTI2018a}
F.~Battisti, S.~Baldoni, M.~Brizzi, and M.~Carli.
\newblock A feature-based approach for saliency estimation of omni-directional
  images.
\newblock {\em Signal Processing: Image Communication}, 69:53 -- 59, 2018.

\bibitem{borji2012boosting}
A.~Borji.
\newblock Boosting bottom-up and top-down visual features for saliency
  estimation.
\newblock In {\em 2012 IEEE Conference on Computer Vision and Pattern
  Recognition (CVPR)}, pages 438--445, June 2012.

\bibitem{boularias2011relative}
A.~Boularias, J.~Kober, and J.~Peters.
\newblock Relative entropy inverse reinforcement learning.
\newblock In {\em Proceedings of the Fourteenth International Conference on
  Artificial Intelligence and Statistics}, pages 182--189, 2011.

\bibitem{chen2016infogan}
X.~Chen, Y.~Duan, R.~Houthooft, J.~Schulman, I.~Sutskever, and P.~Abbeel.
\newblock Infogan: Interpretable representation learning by information
  maximizing generative adversarial nets.
\newblock In {\em Advances in neural information processing systems (NIPS)},
  pages 2172--2180, 2016.

\bibitem{cheng2015global}
M.-M. Cheng, N.~J. Mitra, X.~Huang, P.~H. Torr, and S.-M. Hu.
\newblock Global contrast based salient region detection.
\newblock {\em IEEE Transactions on Pattern Analysis and Machine Intelligence},
  37(3):569--582, 2015.

\bibitem{Corbillon2017360Degree}
X.~Corbillon, F.~De~Simone, and G.~Simon.
\newblock 360 degree video head movement dataset.
\newblock In {\em Proceedings of the 8th ACM on Multimedia Systems Conference
  (MMSys'17)}, pages 199--204, 2017.

\bibitem{cornia2016a}
M.~Cornia, L.~Baraldi, G.~Serra, and R.~Cucchiara.
\newblock A deep multi-level network for saliency prediction.
\newblock In {\em 2016 23rd International Conference on Pattern Recognition
  (ICPR)}, pages 3488--3493, Dec 2016.

\bibitem{David2018a}
E.~J. David, J.~Guti{\'e}rrez, A.~Coutrot, M.~P. Da~Silva, and P.~L. Callet.
\newblock A dataset of head and eye movements for 360\&deg; videos.
\newblock In {\em Proceedings of the 9th ACM Multimedia Systems Conference
  (MMSys '18)}, pages 432--437, 2018.

\bibitem{de2017look}
A.~{De Abreu}, C.~{Ozcinar}, and A.~{Smolic}.
\newblock Look around you: Saliency maps for omnidirectional images in vr
  applications.
\newblock In {\em 2017 Ninth International Conference on Quality of Multimedia
  Experience (QoMEX)}, pages 1--6, May 2017.

\bibitem{de2017video}
Y.~S. de~la Fuente, R.~Skupin, and T.~Schierl.
\newblock Video processing for panoramic streaming using hevc and its scalable
  extensions.
\newblock {\em Multimedia Tools and Applications}, 76(4):5631--5659, 2017.

\bibitem{Gaddam2016tiling}
V.~R. Gaddam, M.~Riegler, R.~Eg, C.~Griwodz, and P.~Halvorsen.
\newblock Tiling in interactive panoramic video: Approaches and evaluation.
\newblock {\em IEEE Transactions on Multimedia}, 18(9):1819--1831, Sep. 2016.

\bibitem{goferman2012context}
S.~Goferman, L.~Zelnik-Manor, and A.~Tal.
\newblock Context-aware saliency detection.
\newblock {\em IEEE Transactions on Pattern Analysis and Machine Intelligence},
  34(10):1915--1926, Oct 2012.

\bibitem{harel2007graph}
J.~Harel, C.~Koch, and P.~Perona.
\newblock Graph-based visual saliency.
\newblock In {\em Advances in neural information processing systems}, pages
  545--552, 2007.

\bibitem{ho2016generative}
J.~Ho and S.~Ermon.
\newblock Generative adversarial imitation learning.
\newblock In {\em Advances in Neural Information Processing Systems (NIPS)},
  pages 4565--4573, 2016.

\bibitem{hu2017head}
B.~{Hu}, I.~{Johnson-Bey}, M.~{Sharma}, and E.~{Niebur}.
\newblock Head movements during visual exploration of natural images in virtual
  reality.
\newblock In {\em 2017 51st Annual Conference on Information Sciences and
  Systems (CISS)}, pages 1--6, March 2017.

\bibitem{Hu2017e}
H.~Hu, Y.~Lin, M.~Liu, H.~Cheng, Y.~Chang, and M.~Sun.
\newblock Deep 360 pilot: Learning a deep agent for piloting through 360
  sports videos.
\newblock In {\em 2017 IEEE Conference on Computer Vision and Pattern
  Recognition (CVPR)}, pages 1396--1405, July 2017.

\bibitem{huang2015salicon}
X.~Huang, C.~Shen, X.~Boix, and Q.~Zhao.
\newblock Salicon: Reducing the semantic gap in saliency prediction by adapting
  deep neural networks.
\newblock In {\em 2015 IEEE International Conference on Computer Vision
  (ICCV)}, pages 262--270, Dec 2015.

\bibitem{Itti1998a}
L.~Itti, C.~Koch, and E.~Niebur.
\newblock A model of saliency-based visual attention for rapid scene analysis.
\newblock {\em IEEE Transactions on Pattern Analysis and Machine Intelligence},
  20(11):1254--1259, Nov 1998.

\bibitem{Iwasaki1986relation}
M.~Iwasaki and H.~Inomata.
\newblock Relation between superficial capillaries and foveal structures in the
  human retina.
\newblock {\em Investigative ophthalmology \& visual science},
  27(12):1698--1705, 1986.

\bibitem{Judd2009learning}
T.~{Judd}, K.~{Ehinger}, F.~{Durand}, and A.~{Torralba}.
\newblock Learning to predict where humans look.
\newblock In {\em 2009 IEEE 12th International Conference on Computer Vision
  (ICCV)}, pages 2106--2113, Sep. 2009.

\bibitem{Kanan2009SUNTS}
C.~Kanan, M.~H. Tong, L.~Zhang, and G.~W. Cottrell.
\newblock Sun: Top-down saliency using natural statistics.
\newblock {\em Visual cognition}, pages 979--1003, 2009.

\bibitem{kingma2014adam}
D.~P. Kingma and J.~Ba.
\newblock Adam: A method for stochastic optimization.
\newblock {\em arXiv preprint arXiv:1412.6980}, 2014.

\bibitem{kmmerer2017understanding}
M.~K??mmerer, T.~S.~A. Wallis, L.~A. Gatys, and M.~Bethge.
\newblock Understanding low- and high-level contributions to fixation
  prediction.
\newblock In {\em 2017 IEEE International Conference on Computer Vision
  (ICCV)}, pages 4799--4808, Oct 2017.

\bibitem{Kruthiventi2015deepfix}
S.~S.~S. Kruthiventi, K.~Ayush, and R.~V. Babu.
\newblock Deepfix: A fully convolutional neural network for predicting human
  eye fixations.
\newblock {\em IEEE Transactions on Image Processing}, 26(9):4446--4456, Sep.
  2017.

\bibitem{LEBRETON2018GBVS360}
P.~Lebreton and A.~Raake.
\newblock Gbvs360, bms360, prosal: Extending existing saliency prediction
  models from 2d to omnidirectional images.
\newblock {\em Signal Processing: Image Communication}, 69:69 -- 78, 2018.

\bibitem{Li2018Bridge}
C.~Li, M.~Xu, X.~Du, and Z.~Wang.
\newblock Bridge the gap between vqa and human behavior on omnidirectional
  video: A large-scale dataset and a deep learning model.
\newblock In {\em Proceedings of the 26th ACM International Conference on
  Multimedia (MM '18)}, pages 932--940, 2018.

\bibitem{li2017infogail}
Y.~Li, J.~Song, and S.~Ermon.
\newblock Infogail: Interpretable imitation learning from visual
  demonstrations.
\newblock In {\em Advances in Neural Information Processing Systems (NIPS)},
  pages 3812--3822, 2017.

\bibitem{LING2018a}
J.~Ling, K.~Zhang, Y.~Zhang, D.~Yang, and Z.~Chen.
\newblock A saliency prediction model on 360 degree images using color
  dictionary based sparse representation.
\newblock {\em Signal Processing: Image Communication}, 69:60 -- 68, 2018.

\bibitem{lo2017360}
W.-C. Lo, C.-L. Fan, J.~Lee, C.-Y. Huang, K.-T. Chen, and C.-H. Hsu.
\newblock 360\&deg; video viewing dataset in head-mounted virtual reality.
\newblock In {\em Proceedings of the 8th ACM on Multimedia Systems Conference
  (MMSys'17)}, pages 211--216, 2017.

\bibitem{Monroy2017salnet360}
R.~Monroy, S.~Lutz, T.~Chalasani, and A.~Smolic.
\newblock Salnet360: Saliency maps for omni-directional images with {CNN}.
\newblock {\em CoRR}, 2017.

\bibitem{ORH}
ORH.

\bibitem{ozcinar2018visual}
C.~Ozcinar and A.~Smolic.
\newblock Visual attention in omnidirectional video for virtual reality
  applications.
\newblock In {\em 2018 IEEE Tenth International Conference on Quality of
  Multimedia Experience (QoMEX)}, pages 1--6, 2018.

\bibitem{pan2017salgan}
J.~Pan, C.~Canton{-}Ferrer, K.~McGuinness, N.~E. O'Connor, J.~Torres,
  E.~Sayrol, and X.~{Gir{\'{o}} i Nieto}.
\newblock Salgan: Visual saliency prediction with generative adversarial
  networks.
\newblock {\em CoRR}, 2017.

\bibitem{pan2016shallow}
J.~Pan, E.~Sayrol, X.~Giro-I-Nieto, K.~McGuinness, and N.~E. OConnor.
\newblock Shallow and deep convolutional networks for saliency prediction.
\newblock In {\em 2016 IEEE Conference on Computer Vision and Pattern
  Recognition (CVPR)}, pages 598--606, June 2016.

\bibitem{Rai2017a}
Y.~Rai, J.~Guti{\'e}rrez, and P.~Le~Callet.
\newblock A dataset of head and eye movements for 360 degree images.
\newblock In {\em Proceedings of the 8th ACM on Multimedia Systems Conference
  (MMSys'17)}, pages 205--210, 2017.

\bibitem{Rajashekar2005StatisticalAA}
U.~Rajashekar, I.~Van Der~Linde, A.~C. Bovik, and L.~K. Cormack.
\newblock Gaffe: A gaze-attentive fixation finding engine.
\newblock {\em IEEE transactions on image processing}, 17(4):564--573, 2008.

\bibitem{Ramanishka2017top}
V.~Ramanishka, A.~Das, J.~Zhang, and K.~Saenko.
\newblock Top-down visual saliency guided by captions.
\newblock In {\em 2017 IEEE Conference on Computer Vision and Pattern
  Recognition (CVPR)}, pages 3135--3144, July 2017.

\bibitem{Salvucci2000Identifying}
D.~D. Salvucci and J.~H. Goldberg.
\newblock Identifying fixations and saccades in eye-tracking protocols.
\newblock In {\em Proceedings of the 2000 Symposium on Eye Tracking Research \&
  Applications (ETRA '00)}, pages 71--78, 2000.

\bibitem{simonyan2013deep}
K.~Simonyan, A.~Vedaldi, and A.~Zisserman.
\newblock Deep inside convolutional networks: Visualising image classification
  models and saliency maps.
\newblock {\em arXiv preprint arXiv:1312.6034}, 2013.

\bibitem{Sitzmann2018saliency}
V.~Sitzmann, A.~Serrano, A.~Pavel, M.~Agrawala, D.~Gutierrez, B.~Masia, and
  G.~Wetzstein.
\newblock Saliency in vr: How do people explore virtual environments?
\newblock {\em IEEE Transactions on Visualization and Computer Graphics},
  24(4):1633--1642, April 2018.

\bibitem{startsev2018360}
M.~Startsev and M.~Dorr.
\newblock 360-aware saliency estimation with conventional image saliency
  predictors.
\newblock {\em Signal Processing: Image Communication}, 2018.

\bibitem{Stengle2016display}
M.~Stengel and M.~Magnor.
\newblock Gaze-contingent computational displays: Boosting perceptual fidelity.
\newblock {\em IEEE Signal Processing Magazine}, 33(5):139--148, Sep. 2016.

\bibitem{tieleman2012lecture}
T.~Tieleman and G.~Hinton.
\newblock Lecture 6.5-rmsprop: Divide the gradient by a running average of its
  recent magnitude.
\newblock {\em COURSERA: Neural networks for machine learning}, 4(2):26--31,
  2012.

\bibitem{upenik2017simple}
E.~{Upenik} and T.~{Ebrahimi}.
\newblock A simple method to obtain visual attention data in head mounted
  virtual reality.
\newblock In {\em 2017 IEEE International Conference on Multimedia Expo
  Workshops (ICMEW)}, pages 73--78, July 2017.

\bibitem{Flickr2018}
F.~VR.

\bibitem{Watkins1992}
C.~J. C.~H. Watkins and P.~Dayan.
\newblock Q-learning.
\newblock {\em Machine Learning}, 8(3):279--292, May 1992.

\bibitem{webVR}
WebVR.

\bibitem{wang2018deep}
J.~S. Wenguan~Wang.
\newblock Deep visual attention prediction.
\newblock {\em IEEE Transactions on Image Processing}, 2018.

\bibitem{xu2018predicting}
M.~Xu, Y.~Song, J.~Wang, M.~Qiao, L.~Huo, and Z.~Wang.
\newblock Predicting head movement in panoramic video: A deep reinforcement
  learning approach.
\newblock {\em IEEE Transactions on Pattern Analysis and Machine Intelligence},
  pages 1--1, 2018.

\bibitem{xu2018gaze}
Y.~Xu, Y.~Dong, J.~Wu, Z.~Sun, Z.~Shi, J.~Yu, and S.~Gao.
\newblock Gaze prediction in dynamic 360immersive videos.
\newblock In {\em 2018 IEEE Conference on Computer Vision and Pattern
  Recognition (CVPR)}, pages 5333--5342, June 2018.

\bibitem{yang2017top}
J.~Yang and M.~Yang.
\newblock Top-down visual saliency via joint crf and dictionary learning.
\newblock {\em IEEE Transactions on Pattern Analysis and Machine Intelligence},
  39(3):576--588, March 2017.

\bibitem{Yu2015a}
M.~Yu, H.~Lakshman, and B.~Girod.
\newblock A framework to evaluate omnidirectional video coding schemes.
\newblock In {\em 2015 IEEE International Symposium on Mixed and Augmented
  Reality}, pages 31--36, Sept 2015.

\bibitem{zhang2016bms}
J.~Zhang and S.~Sclaroff.
\newblock Exploiting surroundedness for saliency detection: A boolean map
  approach.
\newblock {\em IEEE Transactions on Pattern Analysis and Machine Intelligence},
  38(5):889--902, May 2016.

\bibitem{zhang2018saliency}
Z.~Zhang, Y.~Xu, J.~Yu, and S.~Gao.
\newblock Saliency detection in 360 degree videos.
\newblock In {\em 2018 IEEE European Conference on Computer Vision (ECCV)},
  pages 504--520, 2018.

\bibitem{Zhao_2015_CVPR}
R.~Zhao, W.~Ouyang, H.~Li, and X.~Wang.
\newblock Saliency detection by multi-context deep learning.
\newblock In {\em 2015 IEEE Conference on Computer Vision and Pattern
  Recognition (CVPR)}, June 2015.

\bibitem{Zhou2017a}
C.~Zhou, Z.~Li, and Y.~Liu.
\newblock A measurement study of oculus 360 degree video streaming.
\newblock In {\em Proceedings of the 8th ACM on Multimedia Systems Conference
  (MMSys'17)}, pages 27--37, 2017.

\bibitem{ZHU2018the}
Y.~Zhu, G.~Zhai, and X.~Min.
\newblock The prediction of head and eye movement for 360 degree images.
\newblock {\em Signal Processing: Image Communication}, 69:15 -- 25, 2018.

\end{thebibliography}
}

\end{document}